\documentclass[preprint,12pt,authoryear]{elsarticle}

\usepackage{amssymb}

\usepackage{subfigure}
\usepackage{makeidx}  
\usepackage{multirow}
\usepackage{multicol}
\usepackage{footmisc}
\usepackage{booktabs}
\usepackage{rotating}
\usepackage{amsmath,amsfonts,amssymb}
\usepackage{bm}
\usepackage{url}
\usepackage[svgnames,dvipsnames]{xcolor}
\usepackage{cprotect}
\usepackage[english]{babel}
\usepackage{float}
\usepackage{adjustbox}
\usepackage{tabularx}
\usepackage{subfig}
\usepackage[ruled,vlined,linesnumbered,titlenumbered]{algorithm2e}
\usepackage{algcompatible}
\usepackage{times}
\usepackage{soul}
\usepackage{verbatim}

\newcolumntype{P}[1]{>{\centering\arraybackslash}p{#1}}
\newcolumntype{M}[1]{>{\centering\arraybackslash}m{#1}}
\usepackage{enumitem}
\usepackage{marginnote}
\usepackage{soul}
\usepackage{listings}

\journal{}

\newcolumntype{L}[1]{>{\raggedright\let\newline\\\arraybackslash\hspace{0pt}}m{#1}}
\newcolumntype{C}[1]{>{\centering\let\newline\\\arraybackslash\hspace{0pt}}m{#1}}
\newcolumntype{R}[1]{>{\raggedleft\let\newline\\\arraybackslash\hspace{0pt}}m{#1}}

\newcounter{inlineenum}
\renewcommand{\theinlineenum}{\alph{inlineenum}}

\begin{document}

\begin{frontmatter}




\title{Exploiting a Stimuli Encoding Scheme of Spiking Neural Networks for Stream Learning}

\author[label1]{Jesus L. Lobo\corref{cor1}}
\author[label1]{Izaskun Oregi}
\author[label2,label3]{Albert Bifet}
\author[label1,label4,label5]{Javier Del Ser} 

\address[label1]{TECNALIA, 48160 Derio, Spain.}
\address[label2]{T\'{e}l\'{e}com ParisTech, Par\'{\i}s, C201-2 France}
\address[label3]{University of Waikato, Hamilton, New Zealand}
\address[label4]{Basque Center for Applied Mathematics (BCAM), 48009 Bilbao, Spain}
\address[label5]{University of the Basque Country UPV/EHU, 48013 Bilbao, Spain}
\cortext[cor1]{Corresponding author: jesus.lopez@tecnalia.com (Jesus L. Lobo). TECNALIA. P. Tecnologico Bizkaia, Ed. 700, 48160 Derio, Spain. Tl: +34 946 430 50. Fax: +34 901 760 009.}

\begin{abstract}
\small{Stream data processing has gained progressive momentum with the arriving of new stream applications and big data scenarios. One of the most promising techniques in stream learning is the Spiking Neural Network, and some of them use an interesting population encoding scheme to transform the incoming stimuli into spikes. This study sheds lights on the key issue of this encoding scheme, the Gaussian receptive fields, and focuses on applying them as a pre-processing technique to any dataset in order to gain representativeness, and to boost the predictive performance of the stream learning methods. Experiments with synthetic and real data sets are presented, and lead to confirm that our approach can be applied successfully as a general pre-processing technique in many real cases.}
\end{abstract}


\end{frontmatter}


\section{Introduction}\label{intro}

The continuous production of tremendous amount of data in the form of fast streams upsets the traditional view in machine learning, thus giving rise to a new emerging paradigm called stream learning (SL). These streams of data evolve generally over time and may be occasionally affected by a change (concept drift) which impacts on their input data distribution, without following the fundamental hypothesis of stationarity upon which the learning theory is based. Learning in non-stationary environments has attracted much attention in the SL community in recent years due to its importance for many real-life applications, such as financial applications, network monitoring, cybersecurity, sensor networks, social networks analysis, among other Big Data scenarios \citep*{chen2014big}. Learning under non-stationary conditions is especially challenging in Online Learning (OL) scenarios, where some systems may impose stringent restrictions, such as only a single sample is provided to the learning algorithm at every time instant, a very limited processing time, a finite amount of memory, and the necessity of having trained models at every scan of the streams of data.

The impact of SL and concept drift (CD) is particularly relevant in context of Internet of Things (IoT) \citep*{de2016iot}, which sets forth a number of challenges that have to do with the nature of the data and the processes that generate them. Here the stationarity hypothesis is far from being the standard scenario: the data generation processes have a strong spatio-temporal dimension, which needs to be considered during the data modeling process if a SL algorithm claims to perform a reliable knowledge extraction. Moreover, IoT devices can regularly fail (e.g. limited battery life-time, loss of connectivity, failure, aging, overheating, etc.) resulting in a change that may affect data distribution. These changes causes that, predictive models trained over these IoT stream data before the change occurs, become obsolete, and do not adapt suitably to the new emerging distribution. Any learning algorithm that is going to be used in such a scenario should be able to cope with CD in evolving environments.

Many of the traditional ML algorithms need to be retrained if they are used in a changing environment, and they fail to scale properly. For this reason there is a pressing need for new algorithms that adapt to changes as fast as possible, while providing good performance scores. A large number of algorithms have been developed to deal eather with CD adaptation \citep*{gama2014survey,webb2016characterizing} and/or CD detection \citep*{barros2018large}. Some of these adaptation techniques rely on artificial neural networks (ANNs), such as Multilayer Perceptron \citep*{minku2012ddd,polikar2001learn++}, which are a biologically inspired paradigm that mimics the process that brain acquires and processes sensory information. Considered as the third generation of ANNs, Spiking Neural Networks (SNNs) are one of the most biologically plausible approaches \citep*{gerstner2002spiking,kasabov2018time} thanks to their neuron model and their realistic brain-like information processing, which eases their implementation on super-fast and reliable hardware architectures. Especially in SL, some SNNs (e.g. evolving SNNs) are found as a reputed approach for their ability to learn continuously and incrementally, which account for their adaptability to non-stationary and evolving scenarios. It is known that brains do not deal with real numbers but with timed spikes, and a key aspect of the SNNs architecture is how information is encoded with such spikes. Gaussian Receptive Fields (GRFs) population encoding scheme constitutes a biologically plausible and well studied method for representing real-valued parameters \citep*{bohte2002error}. However, there is a lack of research on how the application of a GRFs population encoding scheme to a dataset can improve by itself the predictive performance of a dataset, without being a mere encoding module of a SNN, and becoming a relevant pre-processing technique for the SL field. It is here where we find an interesting research gap and challenge to be tackled in this study. 

Data pre-processing (DP) has become essential in current knowledge discovery scenarios (dominated by increasingly large datasets like in SL), which aims at getting more precise learning process, among others goals \citep*{ramirez2017survey}. Concretely, space transformations generate a whole new set of features by combining or transforming the original ones. Most of the space transformation approaches focus on reducing the number of dimensions. However, this study proposes to increase them by applying a GRFs population encoding scheme, which will turn into a higher representativeness of the input data, and a predictive performance improvement balanced with the amount of time spent on the computations. As recent studies of DP \citep*{garcia2016big,ramirez2017survey} encourage researchers to develop new DP techniques for data streams, we have decided to consider it as a general pre-processing technique that can be applied to any SL method of the literature. This approach will be tested with many of the current SL methods and datasets in the state of the art, as we will show in next sections. 

The study is organized as follows: first, Section \ref{snns} provides a general introduction to the GRFs population encoding scheme in SNNs. Section \ref{impact_stimuli} presents an insight about the impact of the GRFs population encoding parameters on the stream data representation. Section \ref{approach} delves into a detailed description of the proposed approach, while Section \ref{exps} presents the experiments. Sections \ref{res} and \ref{disc} show and discuss the obtained results from such experiments respectively. And finally, Section \ref{conc} draws conclusions and proposes future research lines related to this study.

\section{Spiking Neural Networks and Stimuli Encoding Schemes}\label{snns}

\begin{figure}[H]
	\centering
	\vspace{-0.1cm}
	\includegraphics[width=0.8\columnwidth]{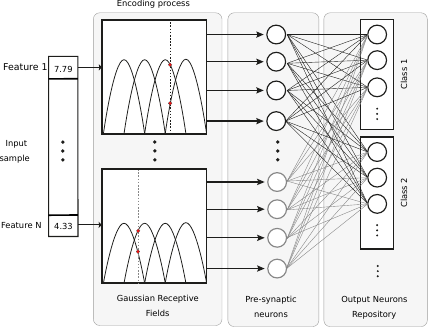}
	\caption{Architecture of an evolving SNN \citep*{kasabov2007evolving}.}
	\label{eSNN_arch}
	\vspace{-0.2cm}
\end{figure}

SNNs have recently attracted much attention in OL due to: 1)  their ability to capture temporal dependence of stream data, 2) they are able to learn continuously and incrementally, 3) they do not need to be retrained in a fast evolving environment, and 4) because they are trained innately in an online manner. Besides, they mimic the process through which the brain acquires and processes sensory information thanks to their biologically plausible neuron models. The use of SNNs in OL allows for a very fast real-time and reducing the computational complexity of the learning process, and this is the case of some approaches like SpikeProp \citep*{bohte2002error}, ReSuMe \citep*{ponulak2005resume}, SpikeTemp \citep*{wang2017spiketemp}, or the recent work presented in \citep*{lobo2018evolving} where the evolving SNNs were modified to fit the OL requirements in a more realistic way. They have been used even as drift detectors in \citep*{lobo2018drift}. Evolving SNNs are a successful type of SNN \citep*{schliebs2013evolving}, where the number of spiking neurons evolves incrementally in time to infer temporal patterns from data. Precisely one of the key ingredient of evolving SNNs is their temporal encoding module (see Figure \ref{eSNN_arch}). The traditional form of patterns which usually consists of real values cannot be used to feed the SNN in a simulation process, and they need to be transformed into temporal patterns (such as events in time or spike trains). Before presenting this pattern to the network, real values of the features of every sample are encoded into spike trains, being a process that aims at generating a new representation of the input stimuli in more dimensions.

SNNs are composed by three layers (see Figure \ref{eSNN_arch}): the first one corresponds to the input data (stimuli). The second layer addresses the encoding process, where the real values of the features of every sample are encoded as trains of spikes by using GRFs. The third layer is the evolving output layer, being a repository of spiking neurons that represent samples and their class labels; they evolve as new samples arrive. Each output neuron is linked to all input neurons through connections whose weights are learned from the samples fed to the network. The second layer is precisely where we focus our encoding approach. 

\begin{figure}[H]
	\centering
	\vspace{-0.1cm}
	\includegraphics[width=0.8\columnwidth]{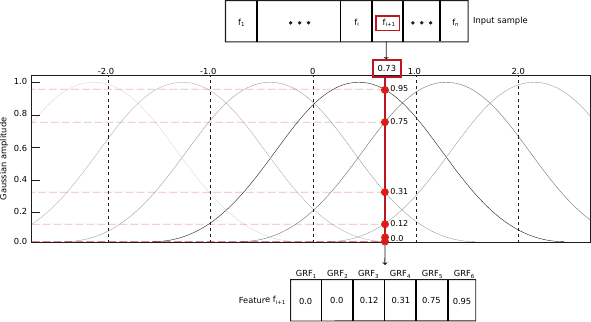}
	\caption{Example of GRFs population encoding based on $6$ Gaussians. For an input value of the feature $f_{i+1}$ ($0.73$, bold straight line), the intersection points with each GRF are computed ($0.95,0.75,0.31,0.12,0.0,0.0$), which are in turn translated into a vector of real values for the feature $f_{i+1}$.}
	\label{GRF}
	\vspace{-0.2cm}
\end{figure}

Rank Order Population encoding \citep*{bohte2002error} is used in the second layer, and it is an extension of the Rank Order encoding introduced in \citep*{thorpe1998rank}. Basically, it allows the mapping of vectors of real values into a sequence of spikes, and it is based on receptive fields which encode the real values by using a collection of Gaussian curves with overlapping sensitivity profiles. Each feature is encoded independently by a number of receptive fields (\textit{n\_GRFs}). GRFs overlap with each other by adopting the shape of a Gaussian function, in all cases covering the whole range of the values of each feature. The parameter \textit{n\_GRFs} may vary depending on the nature of the data at hand, and must be tuned for achieving a good predictive performance of the overall model. Concretely, the center $C_{i}$ and the width $W_{i}$ of each GRF $i$ of the feature $f$ are computed as
\begin{equation}
C_{i}=I_{min}^{n} + \dfrac{2j-3}{2} \left(\frac{I_{max}^{n}-I_{min}^{n}}{\textit{n\_GRFs}-2}\right)
\end{equation}
and
\begin{equation}
W_{i}=\dfrac{1}{\gamma} \left(\dfrac{I_{max}^{n}-I_{min}^{n}}{\textit{n\_GRFs}-2}\right),
\end{equation}
where $n\_GRFs$ is the number of Gaussian receptive fields (equally spaced Gaussian curves), whose value impacts on the amplitude of the input neuron. The range of the $n$-th input feature is $\mathbb{R}[I_{min}^{n},I_{max}^{n}]$. Parameter $\gamma$ (overlap factor) regulates the width of the GRFs, thereby their amount of overlapping. Each feature will be transformed in a real values vector (spikes train), defined as
\begin{equation}
vector_{f}=\exp\left(-\dfrac{(x-C_{i})^{2}}{2 W_{i}^{2}}\right),
\end{equation}
where $x$ is the input value of the feature $f$. Figure \ref{GRF} exemplifies the GRFs encoding process for one of the features ($0.73$) of any given input sample.

\section{Impact of the GRFs Parameters on Stimuli Representation}\label{impact_stimuli}

\begin{figure}[H]
	\centering
	\vspace{-0.1cm}
	\includegraphics[width=0.8\columnwidth]{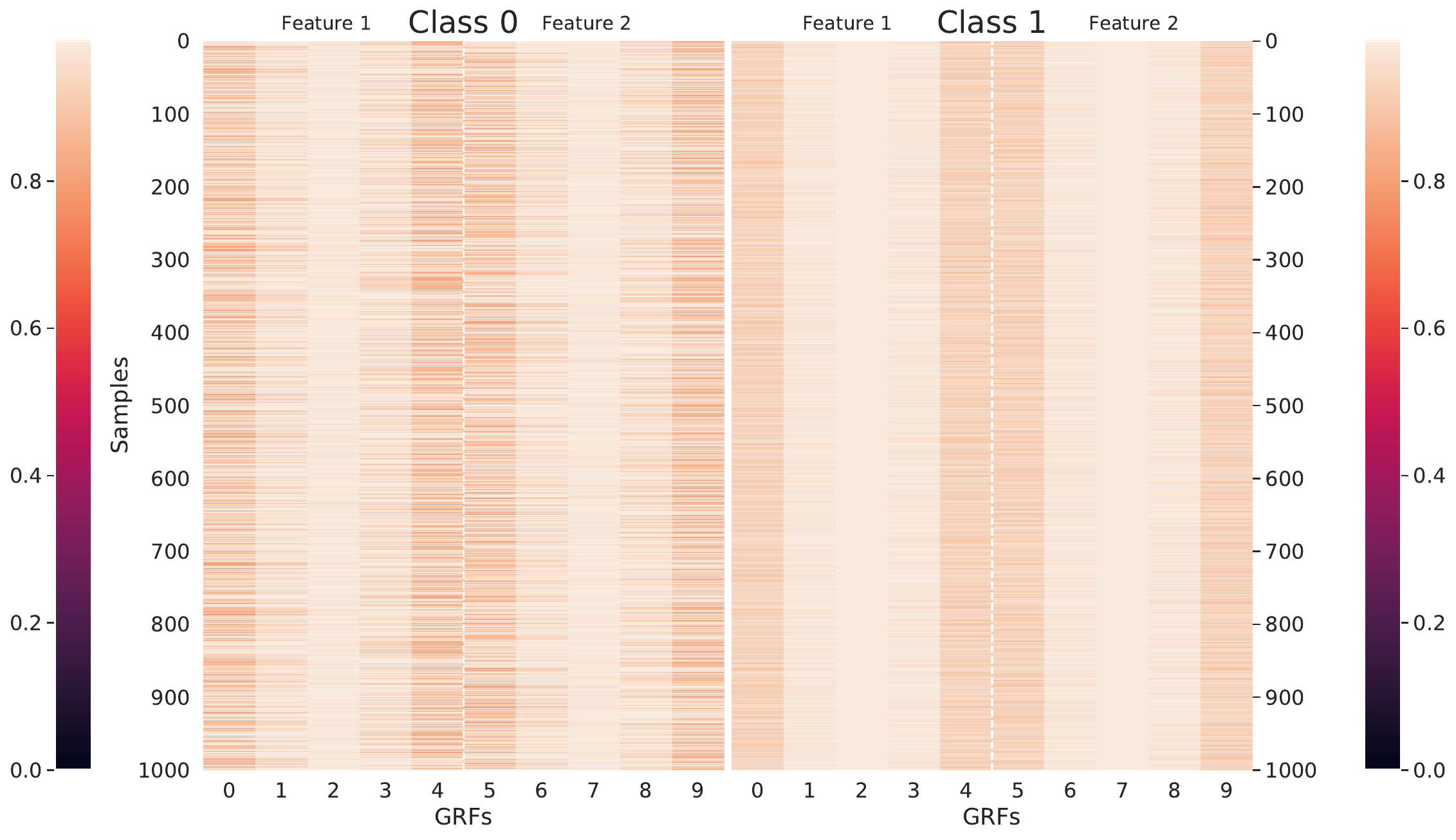}
	\caption{Example of GRFs population encoding: this binary class dataset with $1000$ samples and $2$ features has been encoded with $5$ GRFs (per feature), resulting in an encoded vector of $10$ real values for the samples belonging to the class $0$, and also an encoded vector of $10$ real values for the samples belonging to the class $1$. The color maps correspond to the range of values of the real encoded values.}
	\label{fig:GRF_example}
	\vspace{-0.2cm}
\end{figure}

Before presenting our approach in the next Section, we would like to introduce some insights of the GRFs population encoding scheme when applied to stream data, by showing the impact of the GRFs parameters on the resulting vector of real values. Firstly, and as explained in Section \ref{snns}, the result of the encoding process of a dataset is clearly depicted in Figure \ref{fig:GRF_example}, where it is shown how the parameter \textit{n\_GRFs} impacts on the number of final encoded real values. The more GRFs are used the more cut points with the Gaussian curves are present, and the more real encoded values will be (see lines $11$ and $16$ in Algorithm \ref{alg:GRF_stream_learner}, and Figure \ref{GRF}), which will impact directly in the time processing of each sample, as we will see in Section \ref{disc}.

\begin{figure}[H]
	\centering
	\subfigure[$\textit{gamma}=0.2$]{\includegraphics[width=0.49\columnwidth]{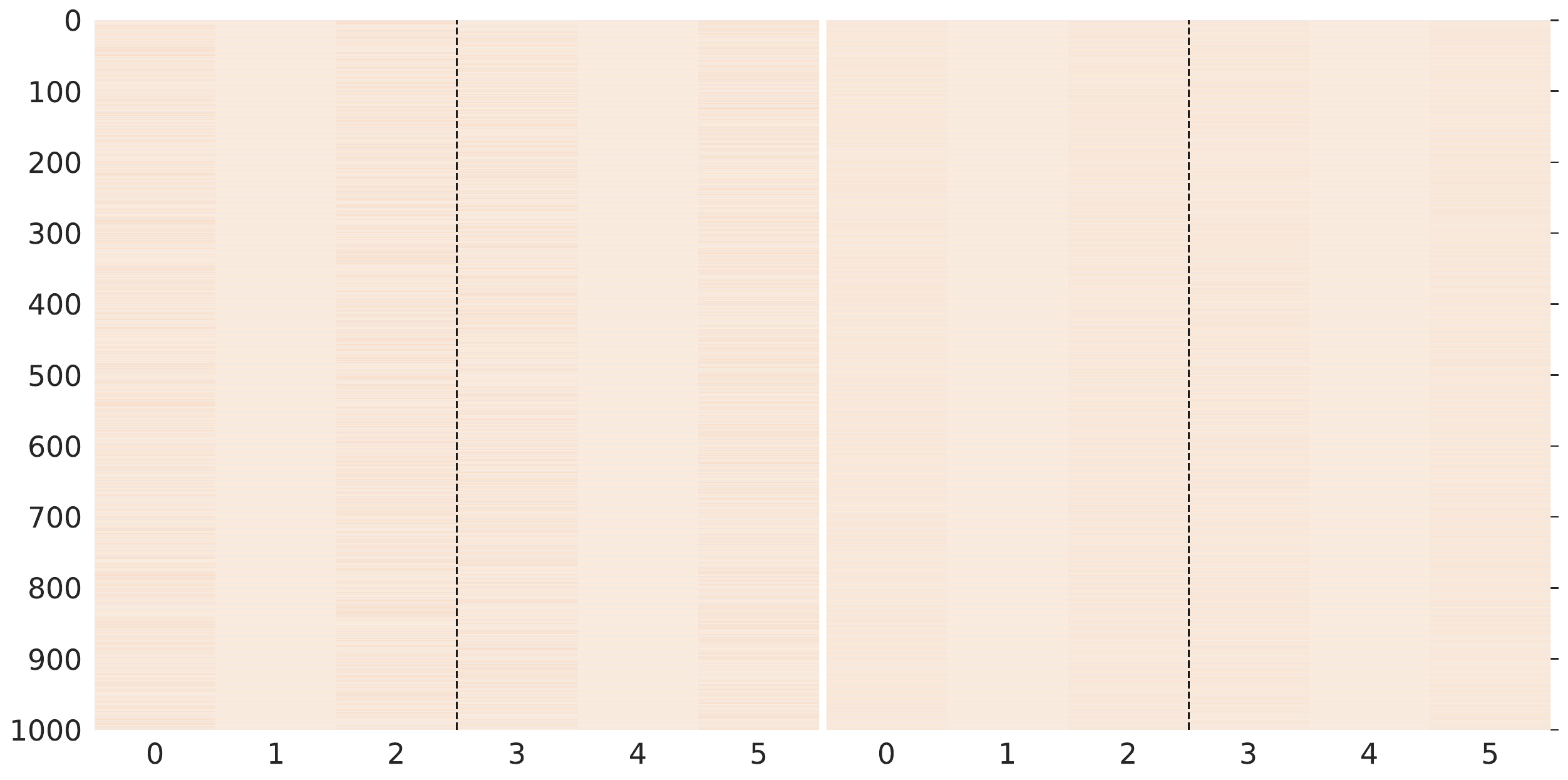}}\hspace{1mm}%
	\subfigure[$\textit{gamma}=1.0$]{\includegraphics[width=0.49\columnwidth]{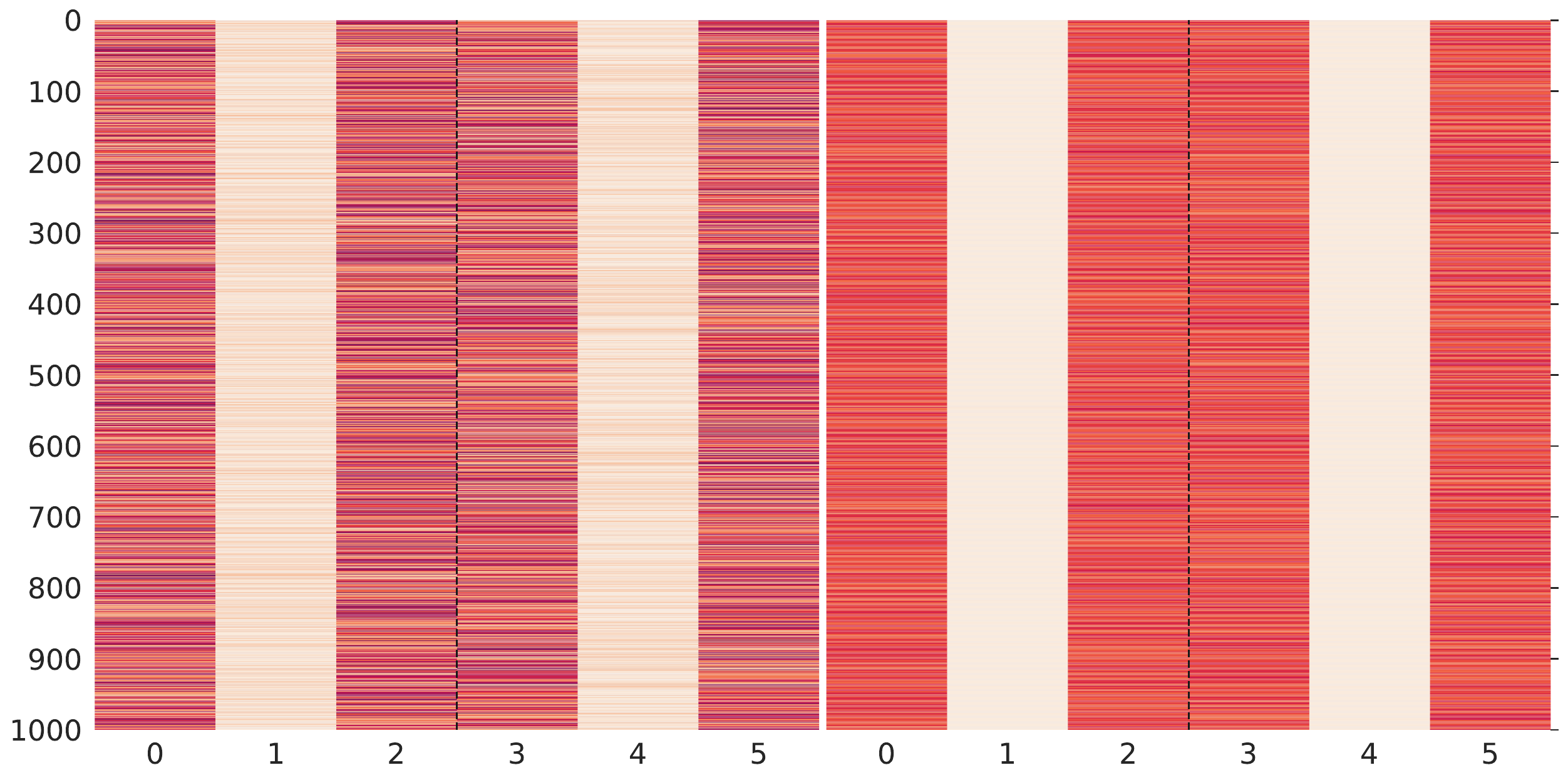}}\hspace{1mm}%
	\subfigure[$\textit{gamma}=2.0$]{\includegraphics[width=0.49\columnwidth]{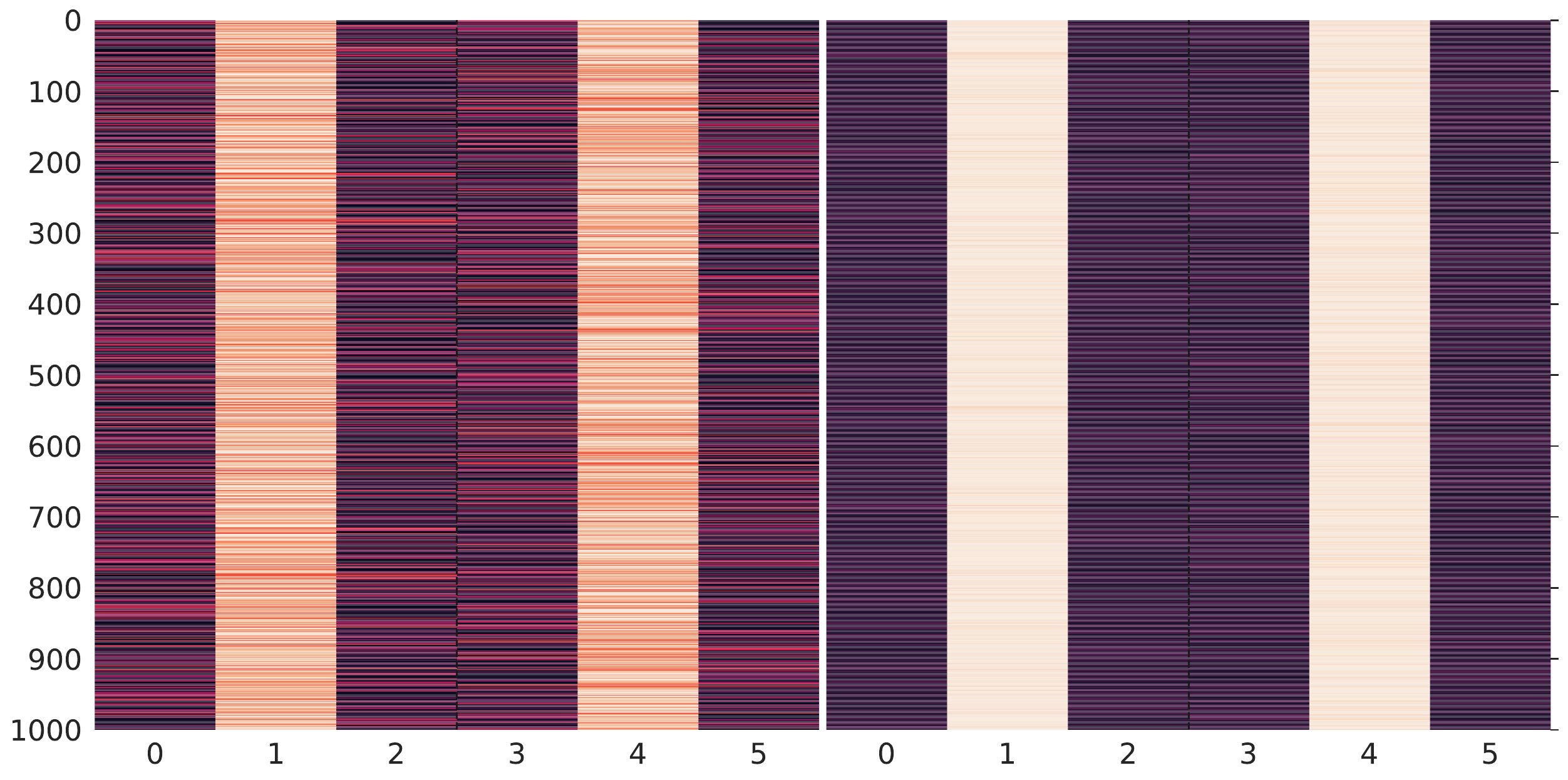}}	
	\caption{Impact of the different values of \textit{gamma} on the real encoded values. They represent an overlap of $2\%$, $10\%$, and $20\%$ between two subsequent GRFs respectively. The parameter \textit{n\_GRFs} is $3$. The characteristics of the dataset and the color map are the same than in Figure \ref{fig:GRF_example}.}
	\label{fig:gamma_impact}
\end{figure}

In Figure \ref{fig:gamma_impact} we can see the impact of the parameter \textit{gamma} on the resulting vector of real values that represent the sample after the GRFs population encoding. The higher value of \textit{gamma} is the lower encoded values are obtained. This makes sense when we see in Figure \ref{GRF} that the overlapping area between Gaussian curves is defined by \textit{gamma} parameter; with a high value of \textit{gamma} we obtain less overlapping and thus the encoded real values will be closer to $0$ (the lower limit of the transformation interval) more frequently. However, with a low value of \textit{gamma}, the overlapping area between Gaussian curves is higher, what provokes that the values of the cut points are closer to the upper limit of the transformation interval. Therefore, when the value of \textit{gamma} is too high or too low (there is no so variety in the values representation, all of them are close to the lower or upper limit of the transformation interval respectively), and then the final encoded vector loses representativeness. The increase in representativeness is precisely the goal of using a GRFs population encoding scheme in this study; by augmenting the representativeness of input data we may increment the predictive power of stream learning methods in many cases, as it will be shown in Sections \ref{res} and \ref{disc}. It is necessary to find the suitable level of overlapping to maximize this predictive performance gain. 

\begin{figure}[H]
	\centering
	\subfigure[$3$ GRFs with a low value of \textit{gamma}]{\includegraphics[width=0.49\columnwidth]{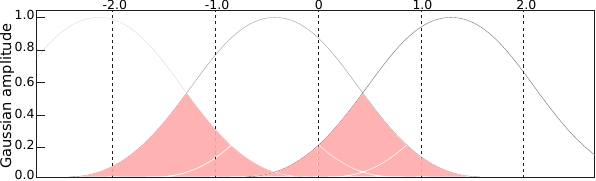}}\hspace{1mm}%
	\subfigure[$20$ GRFs with a high value of \textit{gamma}]{\includegraphics[width=0.49\columnwidth]{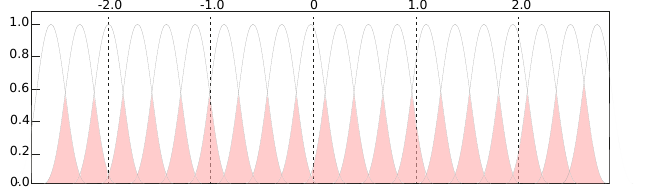}}\hspace{1mm}%
	\caption{Achieving similar overlapping by controlling \textit{gamma} and \textit{n\_GRFs}.}
	\label{fig:overlapping_control}
\end{figure}

\begin{figure}[H]
	\centering
	\subfigure[$\textit{n\_GRFs}=10$]{\includegraphics[width=0.45\columnwidth]{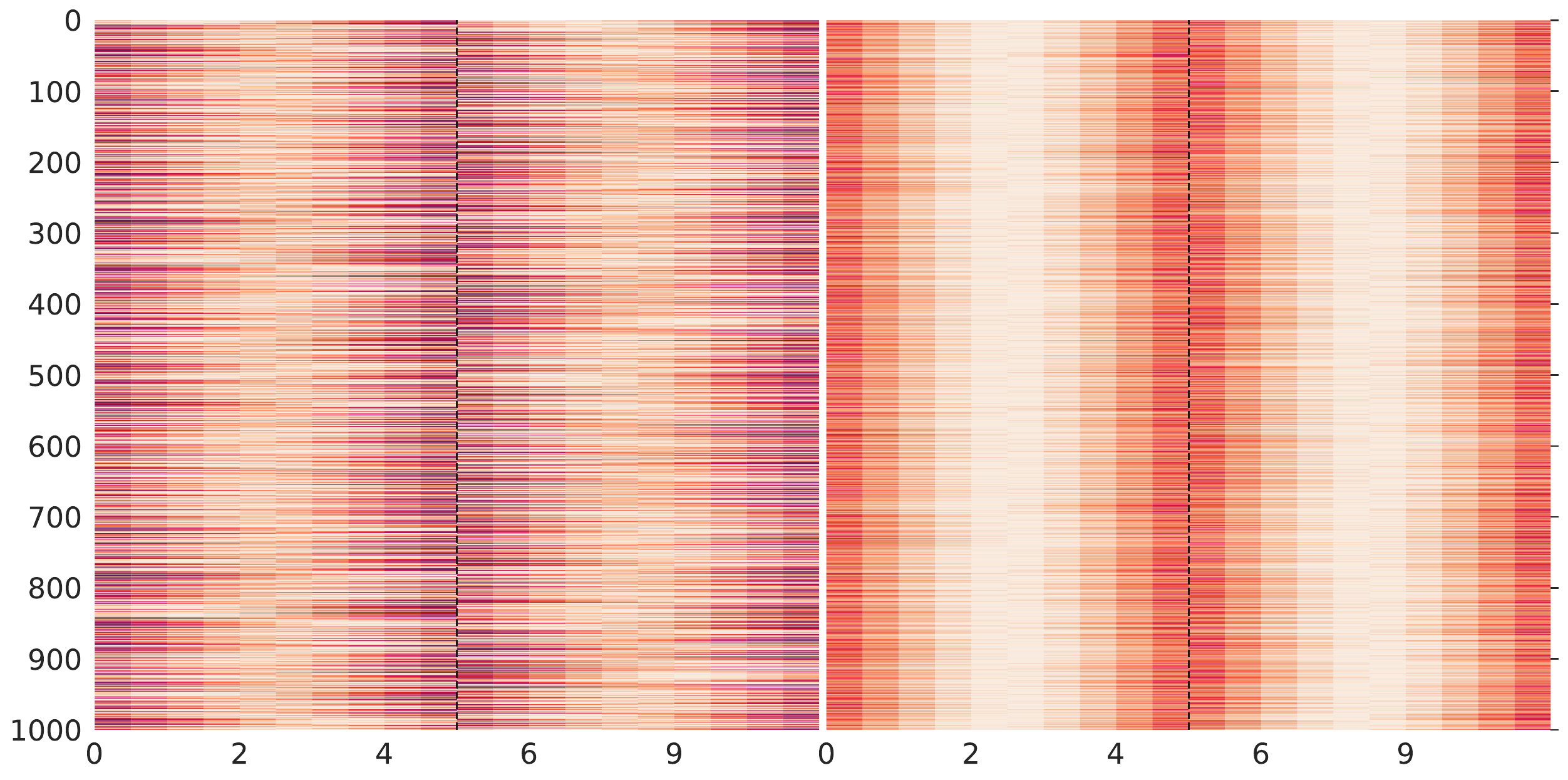}}\hspace{1mm}%
	\subfigure[$\textit{n\_GRFs}=25$]{\includegraphics[width=0.45\columnwidth]{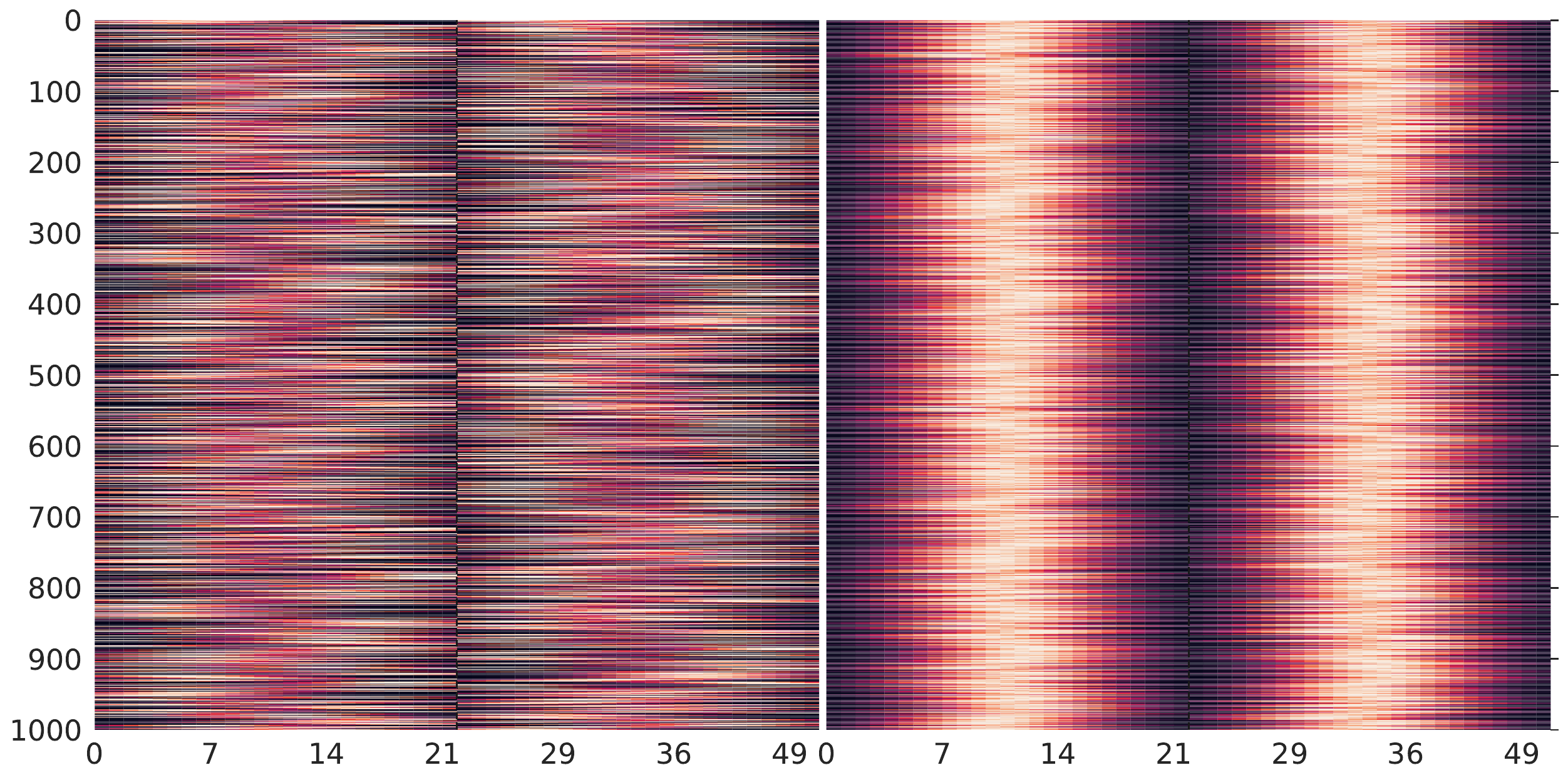}}\hspace{1mm}%
	\subfigure[$\textit{n\_GRFs}=50$]{\includegraphics[width=0.45\columnwidth]{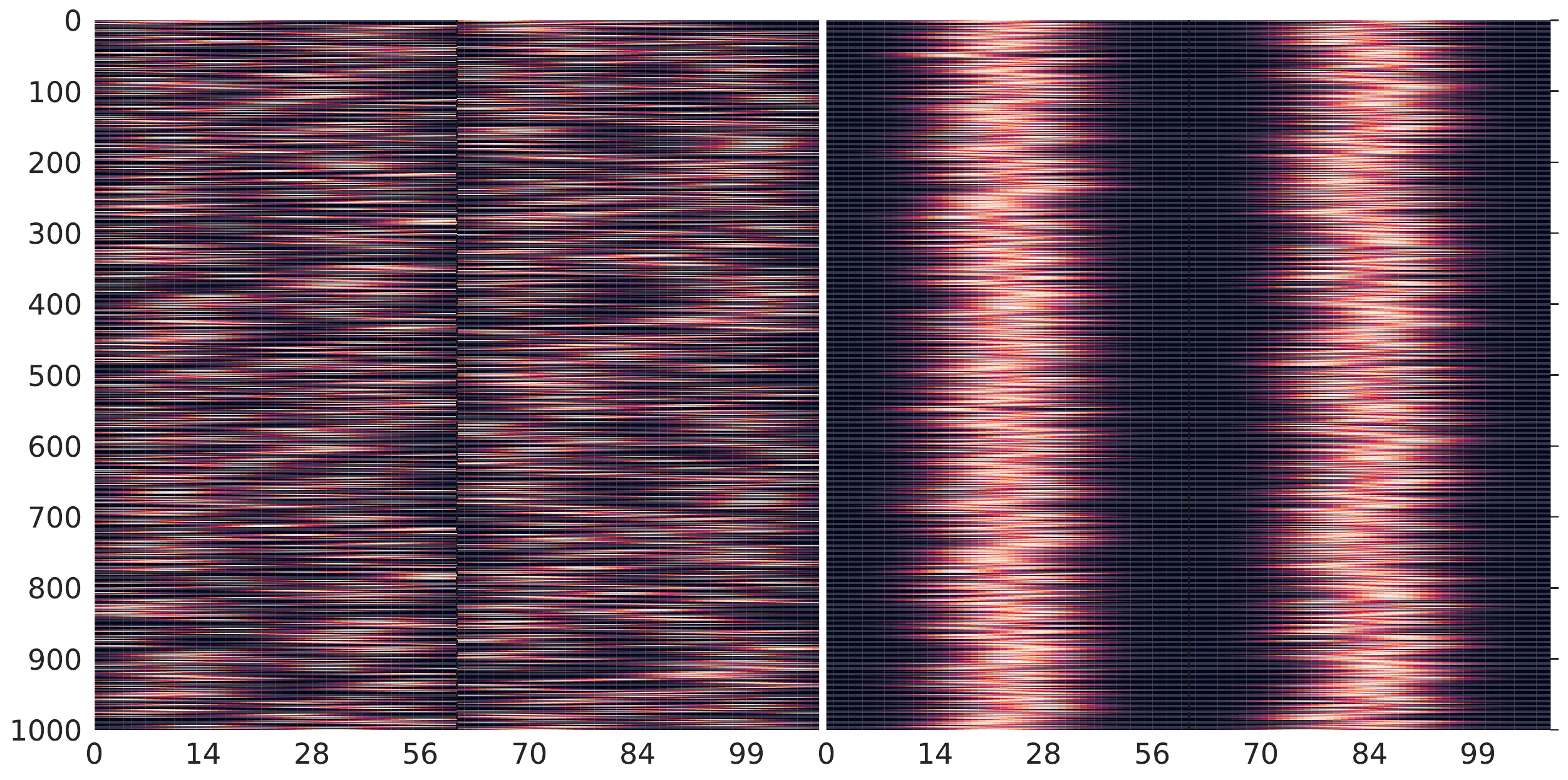}}	
	\caption{Impact of the different values of \textit{n\_GRFs} per feature on the real encoded values. The parameter \textit{gamma} is $0.2$. The characteristics of the dataset and the color map are the same than in Figure \ref{fig:GRF_example}.}
	\label{fig:GRFs_impact}
\end{figure}

\begin{figure}[H]
	\centering
	\subfigure[$\textit{n\_GRFs}=10$]{\includegraphics[width=0.45\columnwidth]{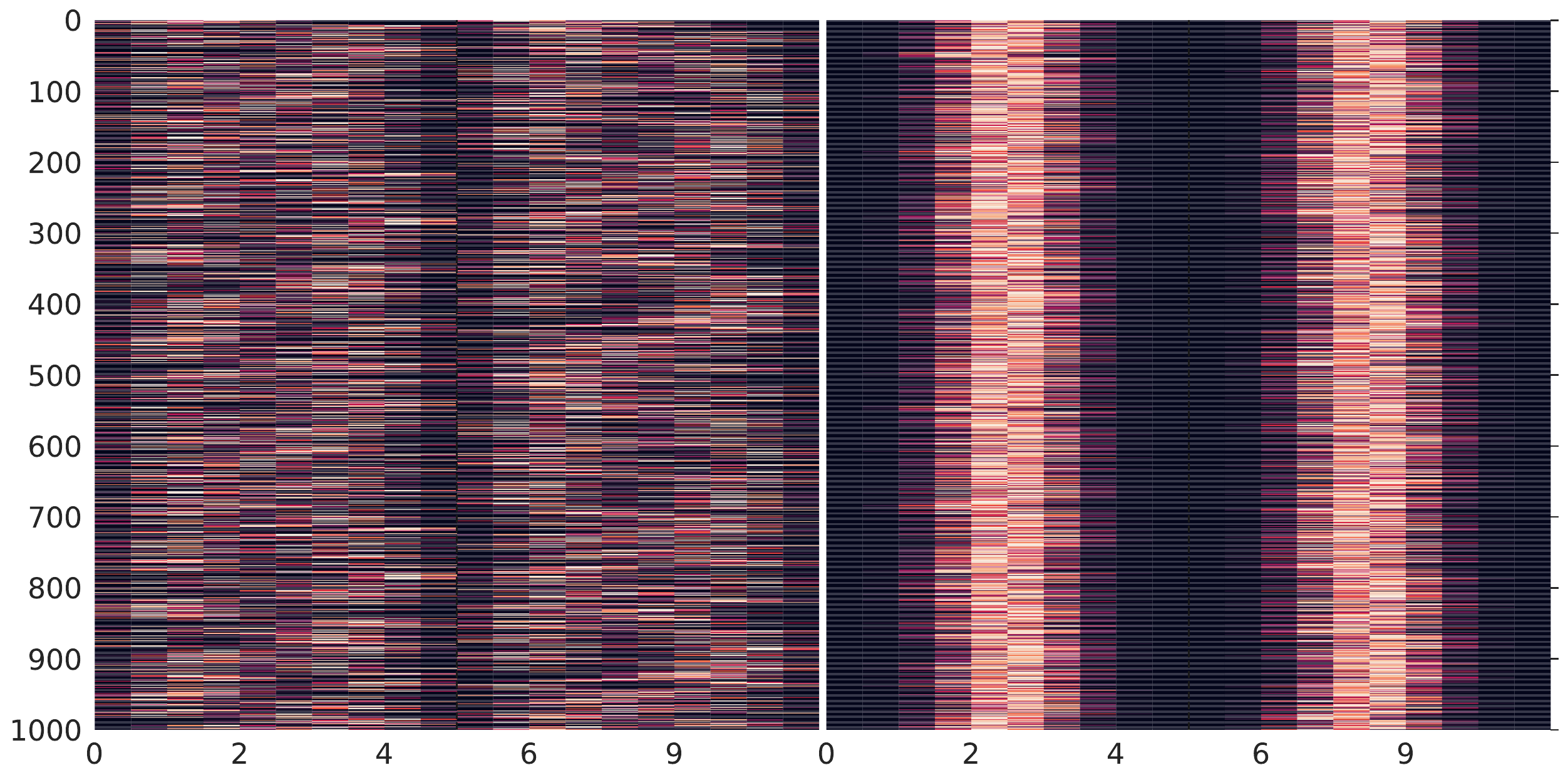}}\hspace{1mm}%
	\subfigure[$\textit{n\_GRFs}=25$]{\includegraphics[width=0.45\columnwidth]{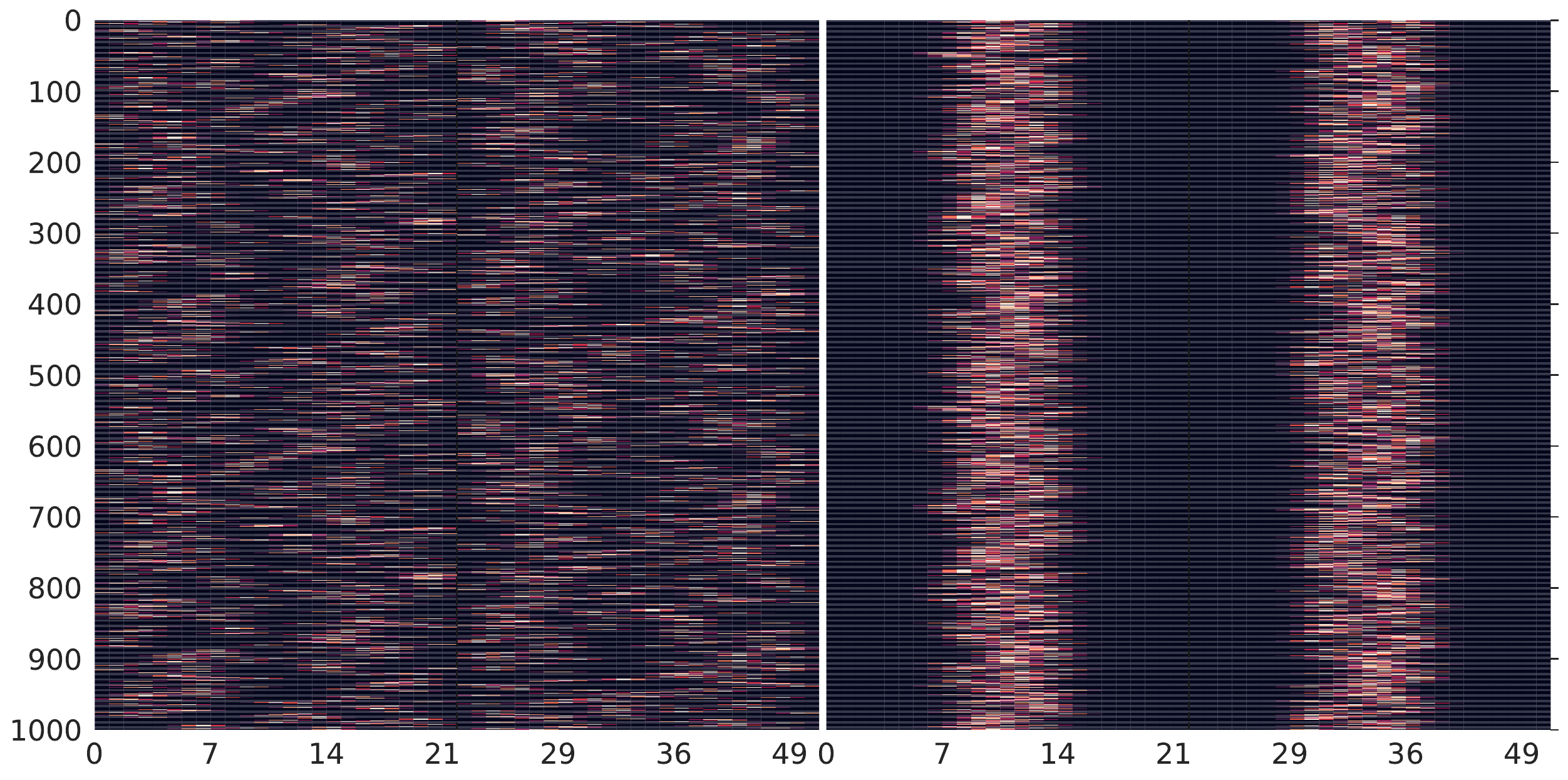}}\hspace{1mm}%
	\subfigure[$\textit{n\_GRFs}=50$]{\includegraphics[width=0.45\columnwidth]{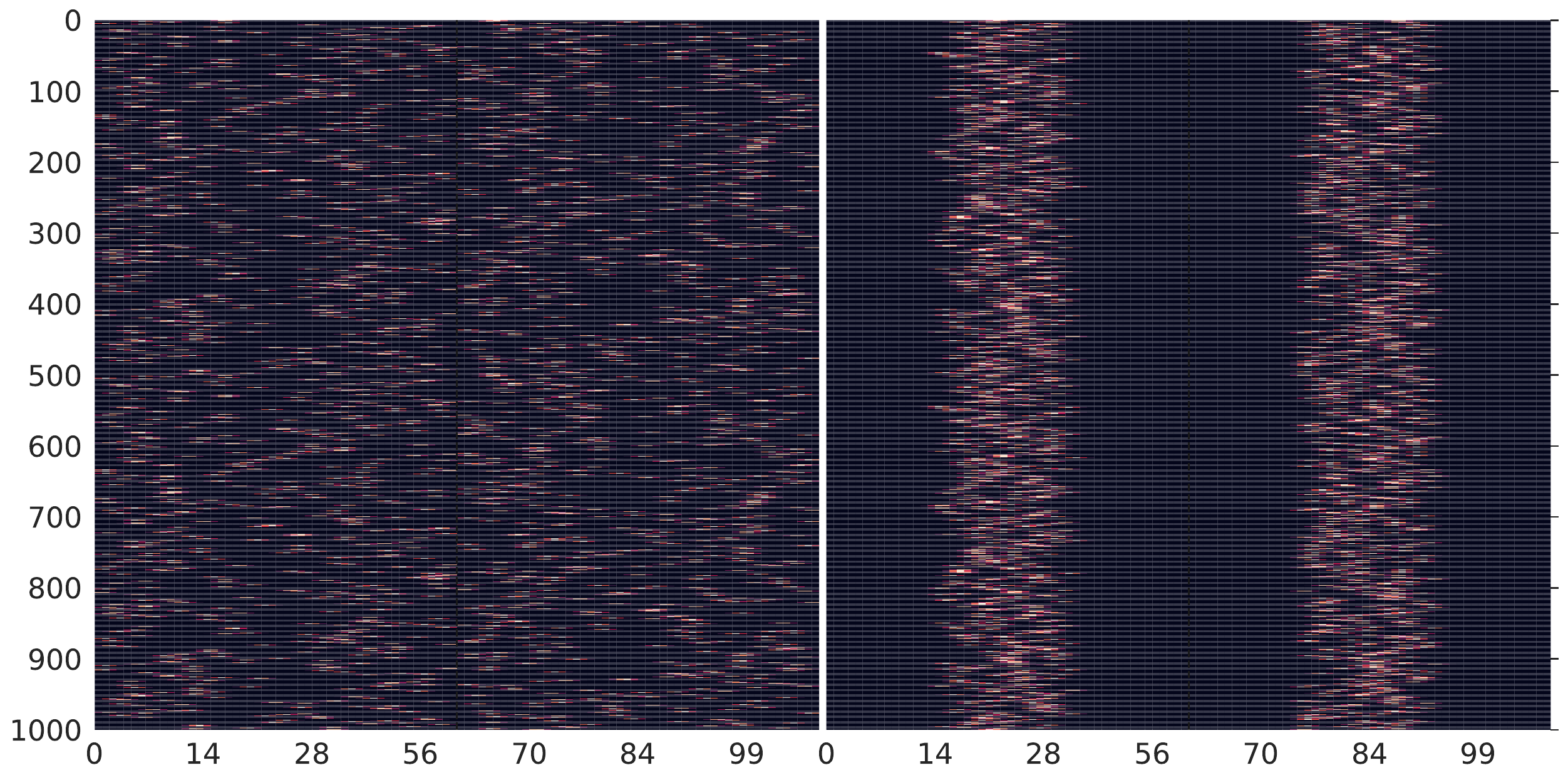}}	
	\caption{Impact of the different values of \textit{n\_GRFs} per feature on the real encoded values. The parameter \textit{gamma} is $1.0$. The characteristics of the dataset and the color map are the same than in Figure \ref{fig:GRF_example}.}
	\label{fig:GRFs_impact2}
\end{figure}

The same situation may happen with \textit{n\_GRFs} parameter, as it is reflected by Figures \ref{fig:GRFs_impact} and \ref{fig:GRFs_impact2}. With a high number of GRFs we obtain a higher number of cut points (see Figure \ref{GRF}) and then a bigger vector of real encoded values. As we see in Figures \ref{fig:GRFs_impact} and \ref{fig:GRFs_impact2}, by fixing the \textit{gamma} parameter and varying the \textit{n\_GRFs} parameter, we also get different levels of representativeness, which lead us to think that there is a trade-off between both parameters of the encoding scheme (see Figure \ref{fig:overlapping_control}). In fact, by combining the suitable values of both parameters we can achieve a similar dataset representation (Figure \ref{fig:GRFs_impact}c and \ref{fig:GRFs_impact2}a). 

Finally, it should be highlighted the fact that features may have different data distributions and representations for each class, and thus the choice of \textit{gamma} and \textit{n\_GRFs} values may affect the features in a different way. This effect can be corroborated in Figures \ref{fig:GRF_example}, \ref{fig:gamma_impact}, \ref{fig:GRFs_impact}, and \ref{fig:GRFs_impact2}.

The goal of this study is to look into the benefits of gaining representativeness of the data, by focusing on its impact on the predictive performance of the stream learning methods. In the next Section we explain our approach from a practical view in stream learning, and after that we set up the experiments to confirm our assumptions.

\section{Proposed Approach} \label{approach}

Our proposed approach consists of applying GRFs population encoding scheme to any SL method, with the objective of achieving an improvement of its predictive performance. This encoding scheme is already used in the literature, but only as a mere encoding module of the evolving SNNs, and there is a lack of research on how its application to any stream learner as a DP technique can be the key to obtain a better predictive performance. 

The application of a GRFs population encoding scheme to any streaming learner is described in Algorithm \ref{alg:GRF_stream_learner}. The aim of this process is to transform the original feature space of a sample (line $6$) into a new set of real values (\textit{vector\_s} of line 7). The transformation process of each feature (line $8$) is based on the creation of a number (\textit{n\_GRFs}) of GRFs (with their own width  and center in lines $9$ and $12$ respectively) and their cut points (line $13$) between the GRFs with the real value that represents each feature (see Figure \ref{GRF}). At the end of the process (line $17$), the sample will be represented by a new set of real values (\textit{vector\_s}). Once the sample has been transformed, a \textit{test-then-train} scheme (see subsection \ref{eval}) is applied, where each sample is firstly used for testing the model before it is used for training (lines $18$ and $24$). During this process, and after the drift detector has been updated with the stream learner prediction (line $19$), a drift may appear (line $20$), and then the drift detector and the stream learner need to be initialized (lines $21$ and $22$). 

\vspace{0.4cm}
\begin{algorithm}[H]
	\DontPrintSemicolon
	\SetAlgoLined
	\SetKwInOut{Input}{Input}
	\SetKwInOut{Output}{Output}
		
	Set GRFs population encoding parameters: $\alpha, \textit{n\_GRFs}$
	
	\textit{Stream\_learner}=[\texttt{KNN}, \texttt{HT}, \texttt{HAT}, \texttt{MNB}, \texttt{GNB}, \texttt{SGD}, \texttt{Perceptron}, \texttt{PA}, \texttt{MLP}] (See subsection \ref{methods})
	
	\textit{Drift\_detector}=ADWIN() (See subsection \ref{dd})
	
	Calculate $I_{max}, I_{min}$ with a portion of the stream data
	
	Perform a warm start of the \textit{stream\_learner} with the same portion of the stream data
	
	\For{every sample \textit{s} in the rest of the stream data}{
		
		Initialize $vector\_{s}$
		
		\For{every feature $f$}{
			
			Calculate $W_{f}$ 
			
			Initialize $vector\_{f}$
			
			\For{every GRF in $n\_GRFs$ $f$}{
			
				Calculate $C_{f}$ 

				Calculate cut points for the GRF

				Concatenate	cut points to $vector\_{f}$
			
			}

			Concatenate	$vector\_{f}$ to $vector\_{s}$
			
		}
				
		Test \textit{stream\_learner} with $vector\_{s}$

		Update \textit{Drift\_detector} with \textit{stream\_learner's} prediction
			
		\If{$\textit{Drift\_detector}==True$}{
			
			Initialize \textit{Stream\_learner}
			
			Initialize \textit{Drift\_detector}
		}
		
		Train \textit{stream\_learner} with $vector\_{s}$
					
	}
	
	\caption{Proposed approach for applying GRFs to any SL method}
	\label{alg:GRF_stream_learner}
\end{algorithm}

\section{Experiments Design} \label{exps} 

An extensive experimental benchmark has been designed in order to confirm the feasibility of applying GRFs population encoding scheme to several SL methods. To achieve this goal, we have selected some of the most utilized stream learners in the literature, and have modified them to include the GRFs population encoding scheme. Finally, we have tested them against their original versions in several synthetic and real datasets following the standards of the SL evaluation. The experiments are composed of $9$ pairs of techniques (original technique vs original technique with GRF population encoding scheme), which are compared in terms of predictive performance and time processing, with a statistical significance test in order to know about the improvement of applying GRFs population encoding scheme to the original versions of the SL methods. Every experiment has been carried out $25$ times. In the next subsections we will detail the set up of these experiments.

\subsection{Framework} \label{frame}

The experiments have been carried out under the \textit{scikit-multiflow} framework \citep*{skmultiflow}, which is implemented in Python given its increasing popularity in the ML community. It is inspired by the most popular open source Java framework for data stream mining, MOA \citep*{MOA-Book-2018}, and it includes a collection of ML algorithms (classification, regression, clustering, outlier detection, CD detection and recommender systems), datasets, and tools and metrics for SL evaluation. It complements \textit{scikit-learn}\footnote{https://scikit-learn.org/stable/}, whose primary focus is batch learning (despite the fact that it also provides researchers with some OL methods) and expands the set of ML tools on this platform.

\subsection{Datasets} \label{data_sets}

The proposed approach is tested with synthetic and real datasets. On the one hand, synthetic datasets are easier to reproduce and identify the data distribution (concept). On the other hand, real-world data allow us to test our approach under real conditions, but without knowing anything about the existence of drifts, their nature (severity, velocity, recurrence, etc.) in case of being present, or the data distribution. 

We have selected four well-known synthetic datasets \citep*{minku2010impact}: \texttt{CIRCLE}, \texttt{LINE}, \texttt{SINEH} and \texttt{SINEV}. Each dataset consists of $2,000$ samples, $2$ normalized and continuous features $\{X_1,X_2\}$, and represents a binary problem. Drift appears at $t=1,000$ in all of them, and we have divided each dataset into $2$ different datasets: one before the drift (first concept) with $1,000$ samples and other one after the drift (second concept) with $1,000$ samples. After that, we have replicated each dataset $50$ times in order to obtain larger datasets more appropriately to serve as test data in a SL process, resulting the following datasets: \textit{circle\_concept1}, \textit{circle\_concept2}, \textit{line\_concept1}, \textit{line\_concept2}, \textit{sine\_concept1}, \textit{sine\_concept2}, \textit{sineH\_concept1}, and \textit{sineH\_concept2}. Resulting ultimately in $8$ different synthetic datasets of $50,000$ samples. In the case of real-world scenarios, we have resorted to $5$ different datasets:

\begin{itemize}[noitemsep,leftmargin=*]
	\item \textit{Weather} dataset \citep*{elwell2011incremental}. The U.S. National Oceanic and Atmospheric Administration has compiled a database with $18,159$ daily weather measurements ($50$ years) from over $9,000$ weather stations all around the world. Data samples are composed of $8$ features (temperature, dew point, sea level pressure, visibility, average wind speed, and other weather related predictors alike). The goal is to infer whether each day was rainy or not. It is available in \textit{scikit-multiflow}.
	\item \textit{Electricity market} dataset \citep*{harries1999splice}. The dataset is based on $45,312$ instances dated from May $1996$ to December $1998$. Each sample refers to a period of $30$ minutes, and has $5$ features (day of week, time stamp, New South Wales electricity demand, Victoria electricity demand, and scheduled electricity transfer between states). The dataset corresponds to a binary problem, and the target identifies the change of the price (up or down) related to a moving average of the last $24$ hours. It is available in \textit{scikit-multiflow}.
	\item \textit{Moving squares} dataset \citep*{losing2016knn}. $4$ equidistantly separated, squared uniform distributions are moving horizontally with constant speed. The direction is inverted whenever the leading square reaches a predefined boundary. Each square represents a different class. The added value of this dataset is the predefined window of $120$ samples before old samples may start to overlap current ones. The dataset contains $200,000$ samples and corresponds to a multi-class classification problem. It is available in \textit{scikit-multiflow}.
	\item \textit{SEA} dataset \citep*{street2001streaming}. It consists of $3$ numerical attributes that vary from $0$ to $10$, where only $2$ of them are relevant to the classification task. A classification function is chosen, among four possible ones. These functions compare the sum of the two relevant attributes with a threshold value, unique for each of the classification functions. Depending on the comparison, the generator will classify an instance as one of the two possible labels. It is available in \textit{scikit-multiflow}, and it has been generated with its \textit{SEAGenerator} method.
	\item \textit{Airlines} dataset \citep*{ikonomovska2011learning}. It contains $539,383$ examples described by $7$ features ($3$ numeric and $4$ nominal). Airlines encapsulates the binary task of predicting whether a given flight will be delayed, given the information of the scheduled departure.		
\end{itemize}

For the sake of time efficiency in the experiments, we have limited the stream data size to the first $50,000$ samples in the case of \textit{Moving Squares} and \textit{Airlines} datasets. Finally, it should be mentioned that synthetic datasets are balanced, whereas real ones are imbalanced data, which will determine the choice of the accuracy performance metric in subsection \ref{eval}.

\subsection{Stream Learning Methods} \label{methods}

The SL methods of this study have been chosen due to their wide use in the state of the art; in fact, all of them are available in the \textit{scikit-multiflow} and \textit{scikit-learn} frameworks. The stream learners are:

\begin{itemize}[noitemsep,leftmargin=*]
	\item \textit{K-Nearest Neighbors} classifier \citep*{dasarathy1991nearest}, labeled as \texttt{KNN}. It works by keeping track of a fixed number (the last max\_window\_size) of training samples. Then, it searches its stored samples and find the closest ones using a selected distance metric.
	\item \textit{Hoeffding Tree} or \textit{Very Fast Decision Tree} \citep*{hulten2001mining}, labeled as \texttt{HT}. It is an incremental decision tree induction algorithm capable of learning from massive data streams, and assumes that the distribution generating samples does not change over time. \texttt{HT} exploits the fact that a small sample can often be enough to choose an optimal splitting attribute. This idea is supported mathematically by the Hoeffding bound, which quantifies the number of samples needed to estimate some statistics within a prescribed precision (e.g. the goodness of an attribute).
	\item \textit{Hoeffding Adaptive Tree} \citep*{bifet2009adaptive}, labeled as \texttt{HAT}. It uses a drift detector, \texttt{ADWIN} \citep*{bifet2007learning}, to monitor performance of branches on the tree and to replace them with new branches when their accuracy decreases if the new branches are more accurate.
	\item \textit{Multinomial Naive Bayes}, labeled as \texttt{MNB}. It implements the Naive Bayes algorithm \citep*{zhang2004optimality} for multinomially distributed data.
	\item \textit{Gaussian Naive Bayes} \citep*{chan1982updating}, labeled as \texttt{GNB}. It updates means and variances in an online manner.
	\item \textit{Stochastic Gradient Descent} classifier \citep*{robbins1985stochastic}, labeled as \texttt{SGD}. It implements regularized linear models with stochastic gradient descent learning. The gradient of the loss is estimated each sample at a time and the model is updated along the way with a decreasing strength schedule (learning rate).
	\item \textit{Perceptron} \citep*{freund1999large}. It is a classification algorithm very similar to \texttt{SGD} classifier. In fact, they share the same underlying implementation, but \textit{Perceptron} uses the perceptron loss function instead of hinge function. By default, \textit{Perceptron} does not require a learning rate, it is not regularized, and it is updated only on mistakes. These characteristics imply that \textit{Perceptron} is slightly faster to train than \texttt{SGD} and that the resulting model is sparser.
	\item \textit{Passive Agressive} classifier \citep*{crammer2006online}, labeled as \texttt{PA}. It is similar to \textit{Perceptron} in that it does not require a learning rate. However, it includes a regularization parameter \textit{C}.
	\item \textit{Multi-layer Perceptron} classifier \citep*{hinton1990connectionist}, labeled as \texttt{MLP}. It trains iteratively since at each time step the partial derivatives of the loss function with respect to the model parameters are computed to update the parameters.
\end{itemize}

\texttt{KNN}, \texttt{HT}, \texttt{HAT} and \texttt{MNB} are available in \textit{scikit-multiflow}, while \texttt{GNB}, \texttt{SDG}, \texttt{Perceptron}, \texttt{PA}, and \texttt{MLP} are available in \textit{scikit-learn} between its options for incremental learning and strategies to scale computationally. Finally, it should be highlighted that all stream learners have been pre-trained with a number of samples before starting the evaluation, in order to enforce a warm start. These pre-training samples are also used to calculate $I_{max}, I_{min}$, and these limits will be used for the rest of the stream data. Other valid option is not to use this pre-training, and update $I_{max}$ and $I_{min}$ values every time a new sample is available.		

\subsection{Drift Detection and Adaptation Mechanisms} \label{dd}

Although the CD detection and adaptation is not the core of this study, it is present in many real streaming processes. For the real-world experiments the drift appearance is unknown, and then a drift detector is needed to identify it as soon as possible, and to trigger an adaptation mechanism (in this work we have opted for an active approach). \texttt{ADWIN} \citep*{bifet2007learning} is a drift detector which maintains a window of variable size containing samples, and automatically grows the window size when no change is apparent, and shrinks it when data changes. It has been selected because can work together with any learning algorithm, it requires low time and memory resources, and provides rigorous guarantees of its performance in the form of limits on the rates of false positives and false negatives \citep*{gonccalves2014comparative}.

Regarding the adaptation, as we do not know anything about the nature of the drifts in the real cases, we can not opt for a suitable adaptation technique beforehand, and we have initialized the learners after drift is detected. Besides, as we have carried out an OL approach in which only one sample is available at each moment, there is no option for storing a window of past samples and performing a forgetting mechanism based on windowing (or other sample storing scheme). The impact of the drift detection and adaptation mechanisms are not the core of this study, and we encourage other researchers to consider other mechanisms. 

\subsection{Streaming Evaluation Methodology} \label{eval}

Evaluation is a fundamental task to know when an approach is outperforming another method only by chance, or when there is a statistical significance to that claim. In the case of SL, the methodology is very specific to consider the fact that not all data can be stored in memory (e.g. in OL only one sample is processed at each time), and that data can follow a non-stationary distribution (drifts may occur at any time). We have followed the evaluation methodology proposed in \citep*{gama2014survey,bifet2015efficient,MOA-Book-2018}, and that recommends to follow these guidelines in several evaluation tasks:

\begin{itemize}[noitemsep,leftmargin=*]
	\item \textit{Error estimation}. We have used an \textit{interleaved test-then-train} scheme, where each sample is firstly used for testing the model before it is used for training, and from this, the accuracy metric is incrementally updated. The model is thus always being tested on samples it has not seen. In our study we have not used a landmark window, and only one sample is used at each time step.
	\item \textit{Performance evaluation measure}. As in real data streams the number of samples for each class may be evolving and changing, we have opted for the Kappa statistic (\textit{K}) \citep*{cohen1960coefficient} because it is a more sensitive measure for quantifying the predictive performance of streaming classifiers:
	\begin{equation}
	\textit{K}=\dfrac{p_{o}-p_{c}}{1-p_{c}}
	\end{equation}
	, where $p_{o}$ is the prequential accuracy of the classifier, and $p_{c}$ is the probability that a chance classifier—one that randomly assigns to each class the same number of samples as the classifier under consideration—makes a correct prediction.
	\item \textit{Statistical significance}. When comparing two classifiers, it is necessary to distinguish whether a classifier is better than another one only by chance, or whether there is a statistical significance to ensure that. The McNemar test \citep*{mcnemar1947note} is a non-parametric test used in SL to assess the differences in the performance of two classifiers. The McNemar statistic (\textit{M}) is given as
	\begin{equation}
	\textit{M}=\dfrac{(a-b)^{2}}{a+b}
	\end{equation}
	, where \textit{a} is the number of samples misclassified by the first classifier and correctly classified by the second one, and \textit{b} is the number of samples misclassified by the second classifier and correctly classified by the first one. The test follows the $\tilde{\chi}^2$ distribution. At $0.95$ confidence it rejects the null hypothesis if $\textit{M}>3.841459$ \citep*{dietterich1998approximate}. Because it is well known that the probability of signaling differences where they do not exist is highly affected by data length, we have computed the McNemar test over a sliding window of $500$ samples. \citep*{gama2013evaluating}. The null hypothesis states that augmenting the representativeness of the incoming data by applying a GRFs population encoding scheme, the predictive performance of the stream learning algorithms will be the same than without applying this scheme. We have to reject the null hypothesis at least more than $50\%$ during the SL process to consider that our approach starts to show significance.
	\item \textit{Performance benchmarking}. As it has been shown in subsection \ref{data_sets}, we have taken into consideration both large synthetic and real datasets.
	\item A \textit{cost measure}. We have opted for measuring the processing time (in seconds) of the stream learner in each dataset. The computer that has carried out the experiments is based on a x86\_64 architecture with $8$ processors Intel(R) Core(TM) i7 at $2.70$GHz, and $32$ DDR$4$ memory running at $2,133$ MHz.
\end{itemize}

\subsection{Parameters Choice} \label{params}

Because the best performance of the stream learners is not the goal of this study, we have not carried out a parameter tuning task thoroughly. We are interested in showing the improvement of the classification performance and its statistical significance when GRFs population encoding is applied to the benchmarking. The most relevant parameters configuration is shown in Tables \ref{my_parameters_1}, \ref{my_parameters_2}, and \ref{my_parameters_3} of the \ref{app:params}.

\section{Results} \label{res}

In this section we present the results of the SL methods when the GRF population encoding scheme is applied as a pre-processing technique to synthetic and real-world datasets. We evaluate the classification performance with the Kappa statistic, and the SL processing time in seconds. Regarding the statistical significance, the McNemar column shows the percentage of the data stream in which the null hypothesis is rejected. The null hypothesis to be tested is that there are no significant differences in terms of predictive performance between the original stream learners and those which use the DP technique based on the application of the GRFs encoding scheme.

Tables \ref{circle_res}, \ref{line_res}, \ref{sine_res}, and \ref{sineH_res} collect the results for the synthetic dataset, whereas Tables \ref{real_res1}, \ref{real_res2}, and \ref{real_res3} present the results for the real datasets. Finally, Table \ref{summ_table} summarizes the results in order to have a clearer view of the impact of our approach for each SL method \citep*{bifet2010fast}, providing the mean of the results for all datasets. 

\begin{table}[H]
	\centering
	\resizebox{\textwidth}{!}{%
		\begin{tabular}{|c|c|c|c|c|c|c|}
			\hline
			\multirow{2}{*}{\textbf{Stream Learners}} & \multicolumn{2}{c|}{\textbf{Kappa}} & \multicolumn{2}{c|}{\textbf{McNemar}} & \multicolumn{2}{c|}{\textbf{Time}} \\ \cline{2-7} 
			& \textbf{circle\_concept\_1} & \textbf{circle\_concept\_2} & \textbf{circle\_concept\_1} & \textbf{circle\_concept\_2} & \textbf{circle\_concept\_1} & \textbf{circle\_concept\_2} \\ \hline
			\texttt{KNN} & $0.362$ & $0.300$ & \multirow{2}{*}{$66.59$\%} & \multirow{2}{*}{$100.00$\%} & $12.82\pm0.97$ & $13.02\pm2.35$ \\ \cline{1-3} \cline{6-7} 
			\textit{\texttt{GRF\_KNN}} & $0.414$ & $0.418$ &  &  & $12.58\pm1.10$ & $13.21\pm2.25$ \\ \hline
			\texttt{HT} & $0.980$ & $0.959$ & \multirow{2}{*}{$56.89$\%} & \multirow{2}{*}{$100.00$\%} & $6.03\pm0.50$ & $7.02\pm2.71$ \\ \cline{1-3} \cline{6-7} 
			\textit{\texttt{GRF\_HT}} & $0.998$ & $0.987$ &  &  & $8.67\pm0.28$ & $9.27\pm2.31$ \\ \hline
			\texttt{HAT} & $0.924$ & $0.881$ & \multirow{2}{*}{$14.79$\%} & \multirow{2}{*}{$98.55$\%} & $12.10\pm0.42$ & $12.58\pm1.35$ \\ \cline{1-3} \cline{6-7} 
			\textit{\texttt{GRF\_HAT}} & $0.932$ & $0.886$ &  &  & $16.30\pm0.51$ & $16.64\pm1.97$ \\ \hline
			\texttt{MNB} & $-0.240$ & $-0.182$ & \multirow{2}{*}{$100.00$\%} & \multirow{2}{*}{$100.00$\%} & $17.23\pm1.01$ & $17.60\pm2.35$ \\ \cline{1-3} \cline{6-7} 
			\textit{\texttt{GRF\_MNB}} & $0.794$ & $0.868$ &  &  & $16.62\pm0.69$ & $17.15\pm2.16$ \\ \hline
			\texttt{GNB} & $0.964$ & $0.946$ & \multirow{2}{*}{$0.00$\%} & \multirow{2}{*}{$100.00$\%} & $14.14\pm0.68$ & $14.60\pm1.84$ \\ \cline{1-3} \cline{6-7} 
			\textit{\texttt{GRF\_GNB}} & $0.966$ & $0.950$ &  &  & $15.59\pm0.71$ & $15.88\pm2.02$ \\ \hline
			\texttt{SGD} & $0.012\pm0.003$ & $-0.104\pm0.005$ & \multirow{2}{*}{$100.00$\%} & \multirow{2}{*}{$100.00$\%} & $11.09\pm0.61$ & $11.22\pm1.42$ \\ \cline{1-3} \cline{6-7} 
			\textit{\texttt{GRF\_SGD}} & $0.923\pm0.001$ & $0.929\pm0.001$ &  &  & $11.77\pm0.50$ & $12.12\pm1.67$ \\ \hline
			\texttt{Perceptron} & $-0.679\pm0.004$ & $-0.691\pm0.002$ & \multirow{2}{*}{$100.00$\%} & \multirow{2}{*}{$100.00$\%} & $8.54\pm0.37$ & $8.79\pm1.02$ \\ \cline{1-3} \cline{6-7} 
			\textit{\texttt{GRF\_Perceptron}} & $0.981\pm0.007$ & $0.931\pm0.008$ &  &  & $10.67\pm0.50$ & $11.80\pm3.71$ \\ \hline
			\texttt{PA} & $-0.528\pm0.003$ & $-0.668\pm0.001$ & \multirow{2}{*}{$100.00$\%} & \multirow{2}{*}{$100.00$\%} & $8.44\pm0.37$ & $8.74\pm1.01$ \\ \cline{1-3} \cline{6-7} 
			\textit{\texttt{GRF\_PA}} & $0.925$ & $0.926$ &  &  & $10.59\pm0.48$ & $11.14\pm1.95$ \\ \hline
			\texttt{MLP} & $0.949\pm0.002$ & $0.952\pm0.003$ & \multirow{2}{*}{$36.09$\%} & \multirow{2}{*}{$93.90$\%} & $34.72\pm1.73$ & $35.48\pm4.36$ \\ \cline{1-3} \cline{6-7} 
			\textit{\texttt{GRF\_MLP}} & $0.965\pm0.001$ & $0.962\pm0.001$ &  &  & $37.68\pm1.55$ & $9.32\pm1.47$ \\ \hline
		\end{tabular}%
	}
	\caption{Results for synthetic datasets \textit{circle\_concept1} and \textit{circle\_concept2}. Kappa statistic column evaluates the classification performance, McNemar column shows the percentage of the data stream in which the null hypothesis is rejected, and Time column reflects the duration of the SL process in seconds. The proposed approaches in the column Stream Learners are in italics.}
	\label{circle_res}
\end{table}

\begin{table}[H]
	\centering
	\resizebox{\textwidth}{!}{%
		\begin{tabular}{|c|c|c|c|c|c|c|}
			\hline
			\multirow{2}{*}{\textbf{Stream Learners}} & \multicolumn{2}{c|}{\textbf{Kappa}} & \multicolumn{2}{c|}{\textbf{McNemar}} & \multicolumn{2}{c|}{\textbf{Time}} \\ \cline{2-7} 
			& \textbf{line\_concept\_1} & \textbf{line\_concept\_2} & \textbf{line\_concept\_1} & \textbf{line\_concept\_2} & \textbf{line\_concept\_1} & \textbf{line\_concept\_2} \\ \hline
			\texttt{KNN} & $0.756$ & $0.778$ & \multirow{2}{*}{$92.59$\%} & \multirow{2}{*}{$100.00$\%} & $10.36\pm0.10$ & $15.03\pm1.98$ \\ \cline{1-3} \cline{6-7} 
			\textit{\texttt{GRF\_KNN}} & $0.644$ & $0.718$ &  &  & $10.79\pm0.07$ & $15.13\pm1.85$ \\ \hline
			\texttt{HT} & $0.997$ & $0.997$ & \multirow{2}{*}{$0.00$\%} & \multirow{2}{*}{$0.34$\%} & $4.67\pm0.03$ & $6.39\pm0.94$ \\ \cline{1-3} \cline{6-7} 
			\textit{\texttt{GRF\_HT}} & $0.997$ & $0.986$ &  &  & $7.51\pm0.05$ & $9.17\pm1.32$ \\ \hline
			\texttt{HAT} & $0.978$ & $0.970$ & \multirow{2}{*}{$30.19$\%} & \multirow{2}{*}{$31.27$\%} & $11.78\pm0.07$ & $14.80\pm4.43$ \\ \cline{1-3} \cline{6-7} 
			\textit{\texttt{GRF\_HAT}} & $0.948$ & $0.953$ &  &  & $16.06\pm2.40$ & $19.15\pm3.83$ \\ \hline
			\texttt{MNB} & $0.594$ & $0.578$ & \multirow{2}{*}{$0.00$\%} & \multirow{2}{*}{$100.00$\%} & $15.36\pm0.16$ & $19.32\pm3.38$ \\ \cline{1-3} \cline{6-7} 
			\textit{\texttt{GRF\_MNB}} & $0.974$ & $0.903$ &  &  & $15.50\pm0.79$ & $19.49\pm4.64$ \\ \hline
			\texttt{GNB} & $0.964$ & $0.948$ & \multirow{2}{*}{$0.00$\%} & \multirow{2}{*}{$0.00$\%} & $12.97\pm0.11$ & $16.20\pm2.74$ \\ \cline{1-3} \cline{6-7} 
			\textit{\texttt{GRF\_GNB}} & $0.962$ & $0.956$ &  &  & $14.30\pm0.50$ & $17.69\pm3.39$ \\ \hline
			\texttt{SGD} & $0.972$ & $0.969$ & \multirow{2}{*}{$3.76$\%} & \multirow{2}{*}{$4.45$\%} & $9.78\pm0.09$ & $12.23\pm2.12$ \\ \cline{1-3} \cline{6-7} 
			\textit{\texttt{GRF\_SGD}} & $0.953\pm0.002$ & $0.952\pm0.001$ &  &  & $10.85\pm0.10$ & $13.13\pm2.36$ \\ \hline
			\texttt{Perceptron} & $0.976\pm0.011$ & $0.978\pm0.012$ & \multirow{2}{*}{$73.65$\%} & \multirow{2}{*}{$89.30$\%} & $7.88\pm0.07$ & $9.46\pm1.69$ \\ \cline{1-3} \cline{6-7} 
			\textit{\texttt{GRF\_Perceptron}} & $0.954\pm0.018$ & $0.940\pm0.003$ &  &  & $9.84\pm0.08$ & $11.88\pm2.11$ \\ \hline
			\texttt{PA} & $0.969\pm0.001$ & $0.970$ & \multirow{2}{*}{$3.88$\%} & \multirow{2}{*}{$42.80$\%} & $7.97\pm0.07$ & $9.29\pm1.68$ \\ \cline{1-3} \cline{6-7} 
			\textit{\texttt{GRF\_PA}} & $0.959$ & $0.958$ &  &  & $9.94\pm0.08$ & $11.77\pm2.09$ \\ \hline
			\texttt{MLP} & $0.988$ & $0.984$ & \multirow{2}{*}{$11.00$\%} & \multirow{2}{*}{$57.45$\%} & $31.96\pm0.37$ & $38.28\pm6.53$ \\ \cline{1-3} \cline{6-7} 
			\textit{\texttt{GRF\_MLP}} & $0.983\pm0.001$ & $0.985$ &  &  & $34.83\pm0.43$ & $42.78\pm9.70$ \\ \hline
		\end{tabular}%
	}
	\caption{Results for synthetic datasets \textit{line\_concept1} and \textit{line\_concept2}.}
	\label{line_res}
\end{table}

\begin{table}[H]
	\centering
	\resizebox{\textwidth}{!}{%
		\begin{tabular}{|c|c|c|c|c|c|c|}
			\hline
			\multirow{2}{*}{\textbf{Stream Learners}} & \multicolumn{2}{c|}{\textbf{Kappa}} & \multicolumn{2}{c|}{\textbf{McNemar}} & \multicolumn{2}{c|}{\textbf{Time}} \\ \cline{2-7} 
			& \textbf{sine\_concept\_1} & \textbf{sine\_concept\_2} & \textbf{sine\_concept\_1} & \textbf{sine\_concept\_2} & \textbf{sine\_concept\_1} & \textbf{sine\_concept\_2} \\ \hline
			\texttt{KNN} & $-0.015$ & $-0.009$ & \multirow{2}{*}{$1.00$\%} & \multirow{2}{*}{$13.92$\%} & $13.93\pm0.10$ & $13.28\pm0.71$ \\ \cline{1-3} \cline{6-7} 
			\textit{\texttt{GRF\_KNN}} & $-0.025$ & $-0.018$ &  &  & $14.17\pm0.09$ & $13.52\pm0.92$ \\ \hline
			\texttt{HT} & $0.152$ & $0.139$ & \multirow{2}{*}{$100.00$\%} & \multirow{2}{*}{$32.99$\%} & $6.19\pm0.07$ & $6.21\pm0.25$ \\ \cline{1-3} \cline{6-7} 
			\textit{\texttt{GRF\_HT}} & $0.173$ & $0.160$ &  &  & $9.41\pm0.11$ & $9.50\pm0.14$ \\ \hline
			\texttt{HAT} & $-0.710$ & $-0.149$ & \multirow{2}{*}{$60.37$\%} & \multirow{2}{*}{$66.43$\%} & $11.21\pm0.16$ & $11.64\pm0.09$ \\ \cline{1-3} \cline{6-7} 
			\textit{\texttt{GRF\_HAT}} & $-0.044$ & $0.037$ &  &  & $16.94\pm0.20$ & $16.77\pm0.17$ \\ \hline
			\texttt{MNB} & $0.013$ & $0.001$ & \multirow{2}{*}{$100.00$\%} & \multirow{2}{*}{$0.00$\%} & $17.36\pm0.33$ & $16.93\pm0.35$ \\ \cline{1-3} \cline{6-7} 
			\textit{\texttt{GRF\_MNB}} & $0.042$ & $0.023$ &  &  & $17.11\pm0.18$ & $16.47\pm0.31$ \\ \hline
			\texttt{GNB} & $0.038$ & $0.022$ & \multirow{2}{*}{$3.10$\%} & \multirow{2}{*}{$37.29$\%} & $14.78\pm0.18$ & $14.16\pm0.32$ \\ \cline{1-3} \cline{6-7} 
			\textit{\texttt{GRF\_GNB}} & $0.033$ & $0.044$ &  &  & $16.06\pm0.25$ & $15.39\pm0.30$ \\ \hline
			\texttt{SGD} & $-0.231\pm0.006$ & $-0.300\pm0.006$ & \multirow{2}{*}{$94.54$\%} & \multirow{2}{*}{$92.71$\%} & $11.14\pm0.22$ & $10.65\pm0.21$ \\ \cline{1-3} \cline{6-7} 
			\textit{\texttt{GRF\_SGD}} & $-0.291\pm0.009$ & $-0.298\pm0.006$ &  &  & $11.89\pm0.21$ & $11.38\pm0.16$ \\ \hline
			\texttt{Perceptron} & $-0.696\pm0.001$ & $-0.781\pm0.003$ & \multirow{2}{*}{$7.24$\%} & \multirow{2}{*}{$25.62$\%} & $8.64\pm0.14$ & $8.29\pm0.11$ \\ \cline{1-3} \cline{6-7} 
			\textit{\texttt{GRF\_Perceptron}} & $-0.731\pm0.011$ & $-0.737\pm0.038$ &  &  & $10.77\pm0.18$ & $10.39\pm0.14$ \\ \hline
			\texttt{PA} & $0.626\pm0.003$ & $-0.658\pm0.001$ & \multirow{2}{*}{$64.29$\%} & \multirow{2}{*}{$57.12$ \%} & $8.46\pm0.14$ & $8.33\pm0.09$ \\ \cline{1-3} \cline{6-7} 
			\textit{\texttt{GRF\_PA}} & $-0.881\pm0.001$ & $-0.915\pm0.001$ &  &  & $10.65\pm0.17$ & $10.44\pm0.15$ \\ \hline
			\texttt{MLP} & $0.040\pm0.004$ & $-0.033\pm0.005$ & \multirow{2}{*}{$26.50$\%} & \multirow{2}{*}{$65.09$\%} & $35.02\pm0.59$ & $34.05\pm0.43$ \\ \cline{1-3} \cline{6-7} 
			\textit{\texttt{GRF\_MLP}} & $0.050\pm0.006$ & $0.049\pm0.011$ &  &  & $38.70\pm0.46$ & $37.65\pm0.50$ \\ \hline
		\end{tabular}%
	}
	\caption{Results for synthetic datasets \textit{sine\_concept1} and \textit{sine\_concept2}.}
	\label{sine_res}
\end{table}

\begin{table}[H]
	\centering
	\resizebox{\textwidth}{!}{%
		\begin{tabular}{|c|c|c|c|c|c|c|}
			\hline
			\multirow{2}{*}{\textbf{Stream Learners}} & \multicolumn{2}{c|}{\textbf{Kappa}} & \multicolumn{2}{c|}{\textbf{McNemar}} & \multicolumn{2}{c|}{\textbf{Time}} \\ \cline{2-7} 
			& \textbf{sineH\_concept\_1} & \textbf{sineH\_concept\_2} & \textbf{sineH\_concept\_1} & \textbf{sineH\_concept\_2} & \textbf{sineH\_concept\_1} & \textbf{sineH\_concept\_2} \\ \hline
			\texttt{KNN} & $0.329$ & $0.270$ & \multirow{2}{*}{$0.00$\%} & \multirow{2}{*}{$11.38$\%} & $10.54\pm0.03$ & $12.80\pm1.49$ \\ \cline{1-3} \cline{6-7} 
			\textit{\texttt{GRF\_KNN}} & $0.304$ & $0.288$ &  &  & $10.90\pm0.06$ & $13.09\pm1.42$ \\ \hline
			\texttt{HT} & $0.685$ & $0.657$ & \multirow{2}{*}{$1.69$\%} & \multirow{2}{*}{$1.56$\%} & $5.19\pm0.02$ & $5.87\pm0.45$ \\ \cline{1-3} \cline{6-7} 
			\textit{\texttt{GRF\_HT}} & $0.845$ & $0.808$ &  &  & $8.93\pm0.07$ & $9.26\pm0.21$ \\ \hline
			\texttt{HAT} & $0.426$ & $0.452$ & \multirow{2}{*}{$100.00$\%} & \multirow{2}{*}{$94.40$\%} & $11.81\pm0.05$ & $12.27\pm0.21$ \\ \cline{1-3} \cline{6-7} 
			\textit{\texttt{GRF\_HAT}} & $0.530$ & $0.539$ &  &  & $16.01\pm0.09$ & $16.98\pm0.47$ \\ \hline
			\texttt{MNB} & $0.053$ & $0.122$ & \multirow{2}{*}{$49.47$\%} & \multirow{2}{*}{$12.76$\%} & $15.96\pm0.02$ & $16.55\pm0.77$ \\ \cline{1-3} \cline{6-7} 
			\textit{\texttt{GRF\_MNB}} & $0.464$ & $0.424$ &  &  & $15.81\pm0.03$ & $16.32\pm0.61$ \\ \hline
			\texttt{GNB} & $0.483$ & $0.43$ & \multirow{2}{*}{$0.00$\%} & \multirow{2}{*}{$19.07$\%} & $13.34\pm0.04$ & $13.84\pm0.56$ \\ \cline{1-3} \cline{6-7} 
			\textit{\texttt{GRF\_GNB}} & $0.485$ & $0.524$ &  &  & $14.58\pm0.06$ & $15.10\pm0.49$ \\ \hline
			\texttt{SGD} & $0.165\pm0.003$ & $0.100\pm0.003$ & \multirow{2}{*}{$14.30$\%} & \multirow{2}{*}{$25.49$\%} & $9.92\pm0.02$ & $10.51\pm0.43$ \\ \cline{1-3} \cline{6-7} 
			\textit{\texttt{GRF\_SGD}} & $0.331\pm0.003$ & $0.319\pm0.002$ &  &  & $11.00\pm0.02$ & $11.48\pm0.30$ \\ \hline
			\texttt{Perceptron} & $0.132\pm0.003$ & $0.068\pm0.001$ & \multirow{2}{*}{$34.83$\%} & \multirow{2}{*}{$68.52$\%} & $8.02\pm0.01$ & $8.38\pm0.19$ \\ \cline{1-3} \cline{6-7} 
			\textit{\texttt{GRF\_Perceptron}} & $0.121\pm0.002$ & $0.076\pm0.003$ &  &  & $10.03\pm0.03$ & $10.55\pm0.22$ \\ \hline
			\texttt{PA} & $0.011$ & $-0.076$ & \multirow{2}{*}{$100.00$\%} & \multirow{2}{*}{$100.00$\%} & $7.87\pm0.01$ & $8.23\pm0.17$ \\ \cline{1-3} \cline{6-7} 
			\textit{\texttt{GRF\_PA}} & $0.106$ & $0.079$ &  &  & $9.96\pm0.02$ & $10.32\pm0.28$ \\ \hline
			\texttt{MLP} & $0.708\pm0.023$ & $0.734\pm0.018$ & \multirow{2}{*}{$3.37$\%} & \multirow{2}{*}{$13.82$\%} & $32.47\pm0.14$ & $33.06\pm0.92$ \\ \cline{1-3} \cline{6-7} 
			\textit{\texttt{GRF\_MLP}} & $0.709\pm0.023$ & $0.777\pm0.020$ &  &  & $35.97\pm0.22$ & $39.78\pm1.03$ \\ \hline
		\end{tabular}%
	}
	\caption{Results for synthetic datasets \textit{sineH\_concept1} and \textit{sineH\_concept2}.}
	\label{sineH_res}
\end{table}

\begin{table}[H]
	\centering
	\resizebox{0.8\textwidth}{!}{%
		\begin{tabular}{|c|c|c|c|c|c|c|}
			\hline
			\multirow{2}{*}{\textbf{Stream Learners}} & \multicolumn{2}{c|}{\textbf{Kappa}} & \multicolumn{2}{c|}{\textbf{McNemar}} & \multicolumn{2}{c|}{\textbf{Time}} \\ \cline{2-7} 
			& \textbf{Weather} & \textbf{Electricity} & \textbf{Weather} & \textbf{Electricity} & \textbf{Weather} & \textbf{Electricity} \\ \hline
			\texttt{KNN} & $0.354$ & $0.610$ & \multirow{2}{*}{$6.10$\%} & \multirow{2}{*}{$44.95$\%} & $4.52\pm0.09$ & $13.40\pm0.68$ \\ \cline{1-3} \cline{6-7} 
			\textit{\texttt{GRF\_KNN}} & $0.364$ & $0.640$ &  &  & $6.15\pm0.10$ & $17.53\pm0.90$ \\ \hline
			\texttt{HT} & $0.367$ & $0.508$ & \multirow{2}{*}{$14.29$\%} & \multirow{2}{*}{$40.00$\%} & $3.31\pm0.05$ & $8.29\pm0.48$ \\ \cline{1-3} \cline{6-7} 
			\textit{\texttt{GRF\_HT}} & $0.402$ & $0.491$ &  &  & $7.82\pm0.16$ & $21.28\pm1.27$ \\ \hline
			\texttt{HAT} & $0.300$ & $0.736$ & \multirow{2}{*}{$20.70$\%} & \multirow{2}{*}{$19.52$\%} & $4.82\pm0.08$ & $13.57\pm3.42$ \\ \cline{1-3} \cline{6-7} 
			\textit{\texttt{GRF\_HAT}} & $0.339$ & $0.758$ &  &  & $10.65\pm0.17$ & $23.26\pm1.80$ \\ \hline
			\texttt{MNB} & $0.000$ & $0.002$ & \multirow{2}{*}{$0.00$\%} & \multirow{2}{*}{$81.12$\%} & $5.76\pm0.12$ & $16.57\pm1.18$ \\ \cline{1-3} \cline{6-7} 
			\textit{\texttt{GRF\_MNB}} & $0.000$ & $0.347$ &  &  & $7.01\pm0.10$ & $18.86\pm1.32$ \\ \hline
			\texttt{GNB} & $0.307$ & $0.378$ & \multirow{2}{*}{$25.98$\%} & \multirow{2}{*}{$65.05$\%} & $4.85\pm0.08$ & $13.57\pm0.97$ \\ \cline{1-3} \cline{6-7} 
			\textit{\texttt{GRF\_GNB}} & $0.366$ & $0.271$ &  &  & $6.63\pm0.09$ & $18.09\pm1.22$ \\ \hline
			\texttt{SGD} & $0.403\pm0.004$ & $0.697\pm0.002$ & \multirow{2}{*}{$6.18$\%} & \multirow{2}{*}{$87.90$\%} & $3.71\pm0.07$ & $10.94\pm0.73$ \\ \cline{1-3} \cline{6-7} 
			\textit{\texttt{GRF\_SGD}} & $0.392\pm0.004$ & $0.796\pm0.002$ &  &  & $5.31\pm0.07$ & $14.51\pm0.91$ \\ \hline
			\texttt{Perceptron} & $0.382\pm0.003$ & $0.742\pm0.002$ & \multirow{2}{*}{$4.68$\%} & \multirow{2}{*}{$61.14$\%} & $2.98\pm0.04$ & $8.51\pm0.59$ \\ \cline{1-3} \cline{6-7} 
			\textit{\texttt{GRF\_Perceptron}} & $0.380\pm0.005$ & $0.801\pm0.002$ &  &  & $4.91\pm0.07$ & $13.09\pm0.90$ \\ \hline
			\texttt{PA} & $0.402\pm0.001$ & $0.752\pm0.001$ & \multirow{2}{*}{$17.79$\%} & \multirow{2}{*}{$72.00$\%} & $3.02\pm0.04$ & $8.26\pm0.59$ \\ \cline{1-3} \cline{6-7} 
			\textit{\texttt{GRF\_PA}} & $0.378\pm0.001$ & $0.814\pm0.001$ &  &  & $4.99\pm0.07$ & $13.05\pm0.89$ \\ \hline
			\texttt{MLP} & $0.405\pm0.004$ & $0.545\pm0.005$ & \multirow{2}{*}{$39.70$\%} & \multirow{2}{*}{$57.97$\%} & $12.12\pm0.22$ & $35.01\pm2.42$ \\ \cline{1-3} \cline{6-7} 
			\textit{\texttt{GRF\_MLP}} & $0.442\pm0.002$ & $0.606\pm0.002$ &  &  & $14.75\pm0.24$ & $48.77\pm3.27$ \\ \hline
		\end{tabular}
	}
	\caption{Results for the real datasets: \textit{Weather} and\textit{ Electricity market}.}
	\label{real_res1}
\end{table}

\begin{table}[H]
	\centering
	\resizebox{0.8\textwidth}{!}{%
		\begin{tabular}{|c|c|c|c|c|c|c|}
			\hline
			\multirow{2}{*}{\textbf{Stream Learners}} & \multicolumn{2}{c|}{\textbf{Kappa}} & \multicolumn{2}{c|}{\textbf{McNemar}} & \multicolumn{2}{c|}{\textbf{Time}} \\ \cline{2-7} 
			& \textbf{Moving sq.} & \textbf{SEA} & \textbf{Moving sq.} & \textbf{SEA} & \textbf{Moving sq.} & \textbf{SEA} \\ \hline
			\texttt{KNN} & $0.993$ & $0.623$ & \multirow{2}{*}{$99.96$\%} & \multirow{2}{*}{$8.30$\%} & $17.54\pm0.92$ & $13.46\pm0.17$ \\ \cline{1-3} \cline{6-7} 
			\textit{\texttt{GRF\_KNN}} & $0.931$ & $0.615$ &  &  & $21.82\pm1.17$ & $15.05\pm0.19$ \\ \hline
			\texttt{HT} & $0.104$ & $0.747$ & \multirow{2}{*}{$71.83$\%} & \multirow{2}{*}{$28.00$\%} & $9.10\pm0.53$ & $5.73\pm0.06$ \\ \cline{1-3} \cline{6-7} 
			\textit{\texttt{GRF\_HT}} & $0.160$ & $0.726$ &  &  & $37.95\pm2.07$ & $9.43\pm0.12$ \\ \hline
			\texttt{HAT} & $0.656$ & $0.640$ & \multirow{2}{*}{$80.02$\%} & \multirow{2}{*}{$46.18$\%} & $14.69\pm0.86$ & $10.26\pm0.13$ \\ \cline{1-3} \cline{6-7} 
			\textit{\texttt{GRF\_HAT}} & $0.407$ & $0.695$ &  &  & $58.19\pm3.35$ & $16.96\pm0.22$ \\ \hline
			\texttt{MNB} & $0.060$ & $0.284$ & \multirow{2}{*}{$57.78$\%} & \multirow{2}{*}{$90.38$\%} & $19.87\pm1.25$ & $14.16\pm0.20$ \\ \cline{1-3} \cline{6-7} 
			\textit{\texttt{GRF\_MNB}} & $0.090$ & $0.419$ &  &  & $21.06\pm1.34$ & $14.22\pm0.22$ \\ \hline
			\texttt{GNB} & $0.102$ & $0.755$ & \multirow{2}{*}{$82.80$\%} & \multirow{2}{*}{$6.96$\%} & $18.63\pm1.23$ & $11.90\pm0.18$ \\ \cline{1-3} \cline{6-7} 
			\textit{\texttt{GRF\_GNB}} & $0.102$ & $0.754$ &  &  & $22.52\pm1.45$ & $13.73\pm0.22$ \\ \hline
			\texttt{SGD} & $0.408\pm0.003$ & $0.566\pm0.004$ & \multirow{2}{*}{$100.00$\%} & \multirow{2}{*}{$23.32$\%} & $31.68\pm1.98$ & $9.22\pm0.13$ \\ \cline{1-3} \cline{6-7} 
			\textit{\texttt{GRF\_SGD}} & $0.770\pm0.001$ & $0.591\pm0.002$ &  &  & $32.28\pm2.09$ & $10.44\pm0.15$ \\ \hline
			\texttt{Perceptron} & $0.338\pm0.002$ & $0.482\pm0.002$ & \multirow{2}{*}{$100.00$\%} & \multirow{2}{*}{$14.92$\%} & $25.33\pm1.73$ & $7.21\pm0.11$ \\ \cline{1-3} \cline{6-7} 
			\textit{\texttt{GRF\_Perceptron}} & $0.795\pm0.001$ & $0.473\pm0.004$ &  &  & $29.43\pm1.94$ & $9.40\pm0.14$ \\ \hline
			\texttt{PA} & $0.378\pm0.003$ & $0.395\pm0.001$ & \multirow{2}{*}{$100.00$\%} & \multirow{2}{*}{$55.24$\%} & $25.05\pm1.68$ & $7.03\pm0.10$ \\ \cline{1-3} \cline{6-7} 
			\textit{\texttt{GRF\_PA}} & $0.815$ & $0.466$ &  &  & $29.61\pm1.95$ & $9.26\pm0.13$ \\ \hline
			\texttt{MLP} & $0.966\pm0.006$ & $0.714\pm0.001$ & \multirow{2}{*}{$22.36$\%} & \multirow{2}{*}{$48.83$\%} & $42.71\pm2.61$ & $28.4\pm0.44$ \\ \cline{1-3} \cline{6-7} 
			\textit{\texttt{GRF\_MLP}} & $0.981\pm0.002$ & $0.742\pm0.002$ &  &  & $47.60\pm2.77$ & $31.93\pm0.55$ \\ \hline
		\end{tabular}
	}
	\caption{Results for the real datasets: \textit{Moving squares} and\textit{ SEA}.}
	\label{real_res2}
\end{table}

\begin{table}[H]
	\centering
	\resizebox{0.5\textwidth}{!}{%
		\begin{tabular}{|c|c|c|c|}
			\hline
			\multirow{2}{*}{\textbf{Stream Learners}} & \textbf{Kappa} & \textbf{McNemar} & \textbf{Time} \\ \cline{2-4} 
			& \multicolumn{3}{c|}{\textbf{Airlines}} \\ \hline
			\texttt{KNN} & $0.092$ & \multirow{2}{*}{$100.00$\%} & $15.90\pm0.62$ \\ \cline{1-2} \cline{4-4} 
			\textit{\texttt{GRF\_KNN}} & $1.000$ &  & $24.28\pm0.69$ \\ \hline
			\texttt{HT} & $1.000$ & \multirow{2}{*}{$1.34$\%} & $11.43\pm0.14$ \\ \cline{1-2} \cline{4-4} 
			\textit{\texttt{GRF\_HT}} & $1.000$ &  & $19.13\pm0.25$ \\ \hline
			\texttt{HAT} & $1.000$ & \multirow{2}{*}{$1.34$\%} & $20.73\pm0.31$ \\ \cline{1-2} \cline{4-4} 
			\textit{\texttt{GRF\_HAT}} & $1.000$ &  & $28.07\pm0.42$ \\ \hline
			\texttt{MNB} & $0.095$ & \multirow{2}{*}{$100.00$\%} & $16.88\pm0.25$ \\ \cline{1-2} \cline{4-4} 
			\textit{\texttt{GRF\_MNB}} & $1.000$ &  & $22.19\pm0.34$ \\ \hline
			\texttt{GNB} & $1.000$ & \multirow{2}{*}{$0.00$\%} & $14.50\pm0.23$ \\ \cline{1-2} \cline{4-4} 
			\textit{\texttt{GRF\_GNB}} & $1.000$ &  & $21.88\pm0.40$ \\ \hline
			\texttt{SGD} & $0.115\pm0.002$ & \multirow{2}{*}{$100.00$ \%} & $10.82\pm0.16$ \\ \cline{1-2} \cline{4-4} 
			\textit{\texttt{GRF\_SGD}} & $0.999$ &  & $17.09\pm0.24$ \\ \hline
			\texttt{Perceptron} & $0.116\pm0.002$ & \multirow{2}{*}{$100.00$\%} & $8.25\pm0.12$ \\ \cline{1-2} \cline{4-4} 
			\textit{\texttt{GRF\_Perceptron}} & $1.000$ &  & $15.76\pm0.23$ \\ \hline
			\texttt{PA} & $0.130$ & \multirow{2}{*}{$100.00$ \%} & $8.17\pm0.12$ \\ \cline{1-2} \cline{4-4} 
			\textit{\texttt{GRF\_PA}} & $1.000$ &  & $15.68\pm0.20$ \\ \hline
			\texttt{MLP} & $0.711\pm0.025$ & \multirow{2}{*}{$93.91$\%} & $34.55\pm0.43$ \\ \cline{1-2} \cline{4-4} 
			\textit{\texttt{GRF\_MLP}} & $1.000$ &  & $48.81\pm1.16$ \\ \hline
		\end{tabular}
	}
	\caption{Results for the real dataset: \textit{Airlines}.}
	\label{real_res3}
\end{table}

\begin{table}[H]
	\centering
	\resizebox{0.45\textwidth}{!}{%
		\begin{tabular}{|c|c|c|c|}
			\hline
			\textbf{Stream Learners} & \textbf{Kappa} & \textbf{McNemar} & \textbf{Time} \\ \hline
			\texttt{KNN} & $0.42$ & \multirow{2}{*}{\textit{$52.42$\%}} & \textit{$12.73$} \\ \cline{1-2} \cline{4-4} 
			\textit{\texttt{GRF\_KNN}} & \textit{$0.49$} &  & \textit{$14.87$} \\ \hline
			\texttt{HT} & $0.60$ & \multirow{2}{*}{$26.92$\%} & $6.76$ \\ \cline{1-2} \cline{4-4} 
			\textit{\texttt{GRF\_HT}} & $0.63$ &  & $14.03$ \\ \hline
			\texttt{HAT} & $0.57$ & \multirow{2}{*}{\textit{$51.84$\%}} & \textit{$12.65$} \\ \cline{1-2} \cline{4-4} 
			\textit{\texttt{GRF\_HAT}} & \textit{$0.59$} &  & \textit{$22.18$} \\ \hline
			\texttt{MNB} & $0.13$ & \multirow{2}{*}{\textit{$54.78$\%}} & \textit{$15.80$} \\ \cline{1-2} \cline{4-4} 
			\textit{\texttt{GRF\_MNB}} & \textit{$0.46$} &  & \textit{$16.73$} \\ \hline
			\texttt{GNB} & $0.53$ & \multirow{2}{*}{$31.10$\%} & $13.44$ \\ \cline{1-2} \cline{4-4} 
			\textit{\texttt{GRF\_GNB}} & $0.53$ &  & $16.03$ \\ \hline
			\texttt{SGD} & $0.33$ & \multirow{2}{*}{\textit{$51.79$\%}} & \textit{$12.00$} \\ \cline{1-2} \cline{4-4} 
			\textit{\texttt{GRF\_SGD}} & \textit{$0.56$} &  & \textit{$13.79$} \\ \hline
			\texttt{Perceptron} & $0.23$ & \multirow{2}{*}{\textit{$60.73$\%}} & \textit{$9.46$} \\ \cline{1-2} \cline{4-4} 
			\textit{\texttt{GRF\_Perceptron}} & \textit{$0.48$} &  & \textit{$12.63$} \\ \hline
			\texttt{PA} & $0.22$ & \multirow{2}{*}{\textit{$68.15$\%}} & \textit{$9.36$} \\ \cline{1-2} \cline{4-4} 
			\textit{\texttt{GRF\_PA}} & \textit{$0.47$} &  & \textit{$12.56$} \\ \hline
			\texttt{MLP} & $0.64$ & \multirow{2}{*}{$46.66$\%} & $32.39$ \\ \cline{1-2} \cline{4-4} 
			\textit{\texttt{GRF\_MLP}} & $0.69$ &  & $35.88$ \\ \hline
		\end{tabular}
	}
	\caption{Summarized results for synthetic and real datasets. It contains the mean of Kappa statistics, the mean of the percentages of the McNemar tests in which the null hypothesis is rejected, and the mean of the processing times (seconds) over all datasets. The results in italics mean that there is the enough significance (more than $50\%$ of the length of the dataset) to confirm that the application of GRFs population encoding shows a positive impact on the predictive performance of the SL methods.}
	\label{summ_table}
\end{table}

\section{Discussion} \label{disc}

\begin{figure}[H]
	\centering
	\subfigure[Sliding window=$100$]{\includegraphics[width=0.48\columnwidth]{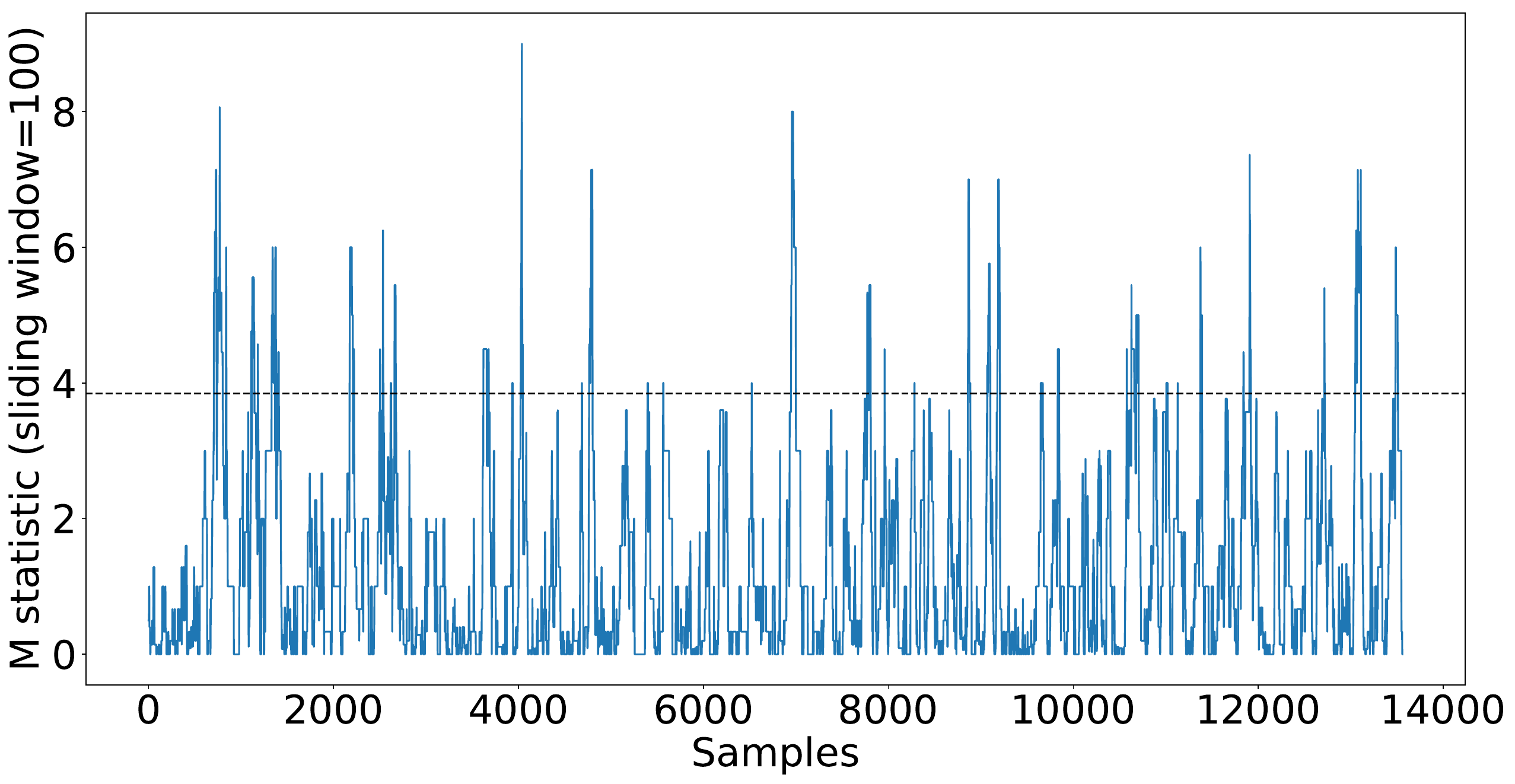}}
	\subfigure[Sliding window=$500$]{\includegraphics[width=0.48\columnwidth]{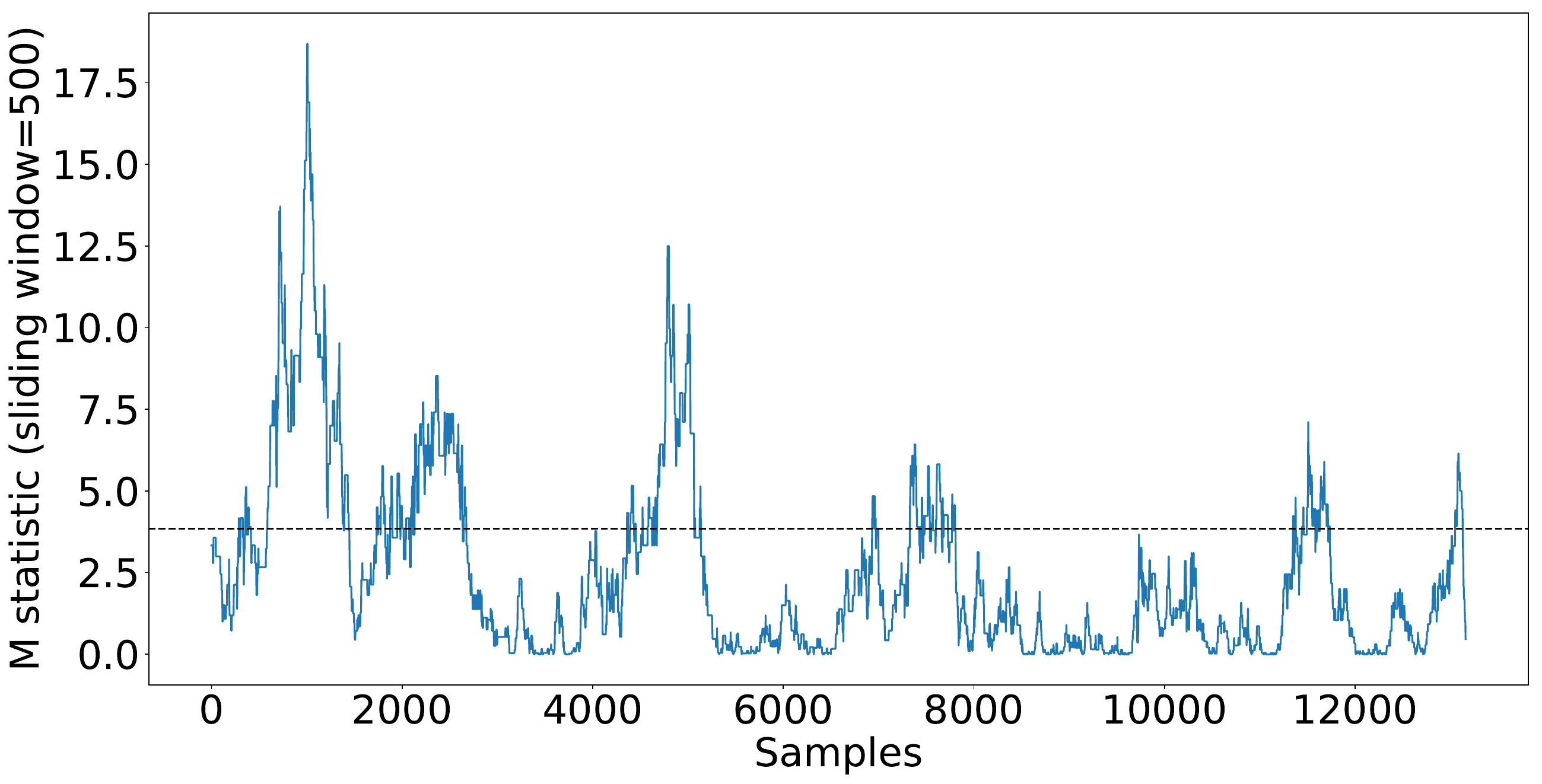}}
	\subfigure[Sliding window=$1000$]{\includegraphics[width=0.48\columnwidth]{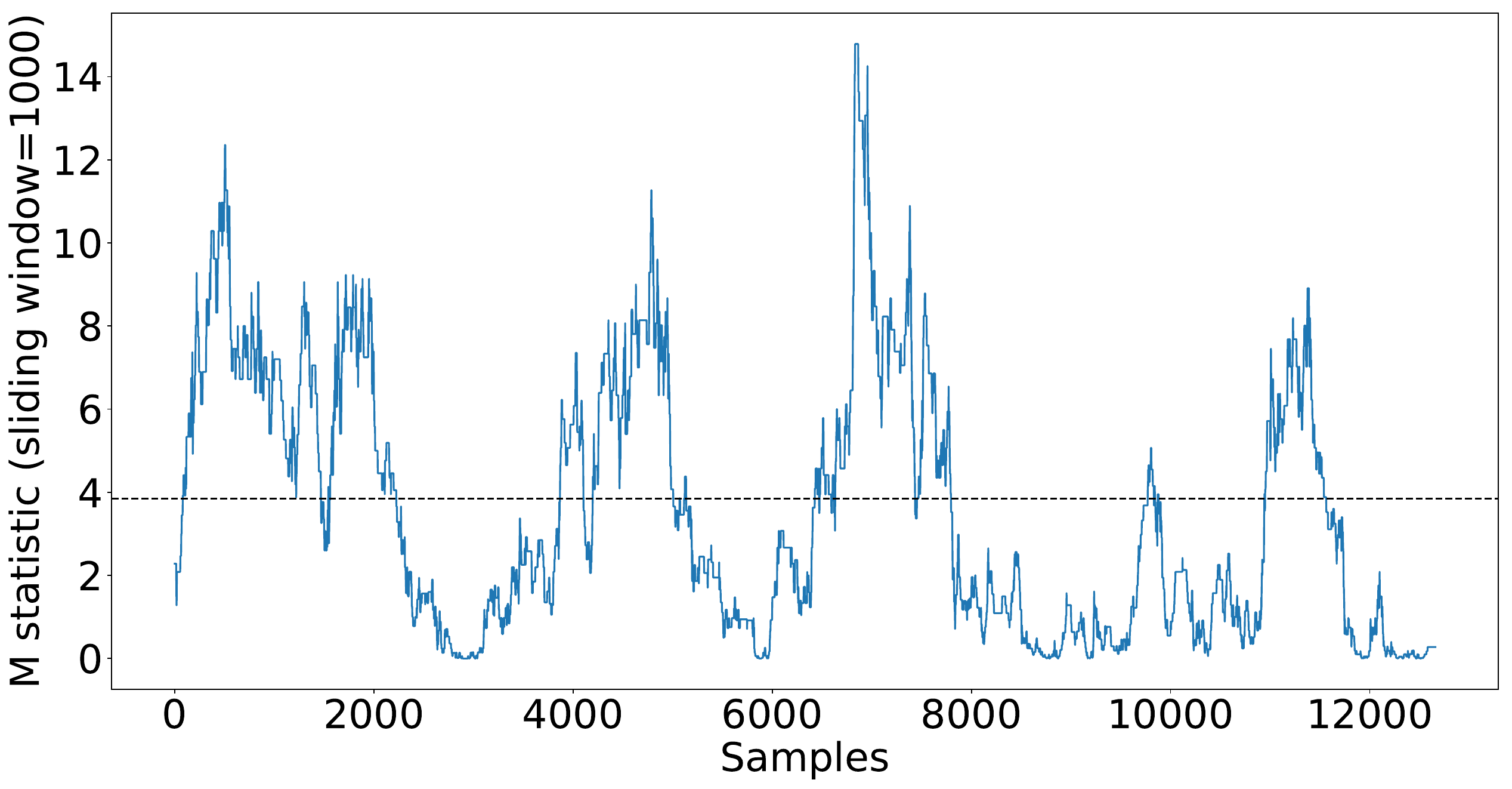}}	
	\caption{The evolution of the McNemar statistic between the \texttt{MLP} stream learner and its version with GRFs population encoding scheme. The dotted line is the threshold for a significance level of 95\%. The null hypothesis is rejected $7.47\%$ (a), $28.81\%$ (b), and $50.23\%$ (c).} \label{fig:mcnemar_SL}
\end{figure}

Regarding the statistical significance of the experiments, in Table \ref{summ_table} we can see where the null hypothesis is rejected more than $50\%$ of the length of the datasets, which means that we may reject the null hypothesis in favor of the hypothesis that the two stream learners have different performance when GRFs population encoding is applied. Concretely, in the case of \texttt{KNN}, \texttt{HAT}, \texttt{MNB}, \texttt{SGD}, \texttt{Perceptron}, and \texttt{PA}, we can confirm that the application of GRFs population encoding has shown a positive impact on the predictive performance of these SL methods. This predictive performance improvement is especially remarkable for \texttt{Perceptron} and \texttt{PA}, where the null hypothesis is rejected in more than $60\%$ of the length of the datasets. For those cases in which the null hypothesis is rejected less than $50\%$, we can also underline that the predictive performance of the SL methods has not been decreased, therefore the application of GRFs population encoding has not impacted negatively on them. For the McNemar test we have chosen a sliding window of $500$ samples. Nevertheless, the experiments in \citep*{gama2013evaluating} pointed out that for different sizes of sliding windows we could obtain different results about the significance of the differences, as Figure \ref{fig:mcnemar_SL} reflects.

Once the statistical significance has been confirmed, we now turn the focus on the predictive performance improvement. In all cases where the statistical significance is evident, the difference between the two stream learners is notable: $0.33$ in \texttt{MNB}, $0.23$ in \texttt{SGD}, $0.25$ in \texttt{Perceptron}, $0.25$ in \texttt{PA}. For \texttt{KNN} and \texttt{HAT} the differences are less considerable ($0.07$ and $0.02$ respectively), but also solid due to the number of repetitions ($25$) of each experiment.

In what refers to the time processing, in all cases the application of GRFs population encoding increases the required time to process the stream data, but with more or less relevancy depending on the stream learner. In the case of \texttt{MNB} we increase the predictive performance in $0.33$ by adding only a $5.88\%$ extra for the processing time, thus we always recommend the application of the DP technique. For \texttt{PA} and \texttt{SGD} the increment in the predictive performance is $0.25$ and $0.23$ respectively, while the extra required processing time is $10.77\%$ and $14.91\%$ respectively; in these cases, we also recommend the application of the DP technique. For \texttt{Perceptron} the increment of the predictive performance is also notable ($0.25$), but we should consider the application of the DP technique carefully because we have to assume a $33.50\%$ extra for the time processing. Finally, for \texttt{KNN} (an increase of $0.07$ in the predictive performance with a $16.81\%$ increment in the processing time) and \texttt{HAT} (an increase of $0.02$ in the predictive performance with a $75.33\%$ increment in the processing time) the application of the DP technique is less recommendable, but we may find some applications fields in which prevails the predictive performance over the processing time. 

This general increment in the processing time of the SL methods makes sense when we consider that GRFs population encoding scheme acts in this study as a DP technique, adding extra calculous to the SL method (more GRFs implies more cut points per feature in the new representation, and then a new larger real-valued vector. See Algorithm \ref{alg:GRF_stream_learner}). Therefore, we will have to tune this parameter good enough to achieve an increment in the predictive performance without drastically penalizing the processing time of the SL algorithm.

\section{Conclusions and Future Work} \label{conc}

This study has elaborated on a new approach for applying a GRFs population encoding scheme to the input data of any stream learning method, augmenting the representativeness of the incoming stimuli, and thus achieving a significant improvement in the predictive performance of the stream learners. A wide variety of synthetic and real datasets have been used for testing a large number of well-known stream learning methods, constituting a complete set of experiments. In order to be sure about the statistical significance of these experiments, McNemar tests have been used to assess the differences in predictive performance of the stream learning methods. This scheme can be seen as a pre-processing technique to be carried out every time a new sample arrives to the stream learning process. Although this scheme increases the processing time at different levels depending on the stream learner method, it has been shown how their predictive performance can boosted, making it worthwhile to apply in those cases where the processing time between samples allows for it. In those cases where the scheme does not reflect a statistical significance, the study has confirmed that the application of this scheme does not harm the predictive performance of the stream learner, only the processing time at different levels. 

Future efforts should be invested on studying how the application of this GRFs population encoding scheme can affect to the drift adaptation, because it could help the stream learner to adapt better to the drift and to recover its predictive performance faster after drift is detected. As it has been mentioned in Section \ref{impact_stimuli}, the choice of the GRFs parameters may have a different impact on the features representation of each class; the possibility of applying different values for the GRFs parameters to each class should be further analyzed. Finally, this pre-processing technique could be also applied to batch learning problems, and we encourage other researcher to apply this technique to batch learning methods with batch datasets.

\section*{Acknowledgements} \label{acknow}

This work was supported by the EU project \textit{iDev40}. This project has received funding from the ECSEL Joint Undertaking (JU) under grant agreement No 783163. The JU receives support from the European Union's Horizon 2020 research and innovation programme and Austria, Germany, Belgium, Italy, Spain, Romania. It has also been supported by the Basque Government (Spain) through the project \textit{VIRTUAL} (KK-2018/00096).

\appendix
\renewcommand*{\thesection}{Appendix \Alph{section}}
\section{Parameter Configuration for the SL Methods} \label{app:params}

\noindent \textbf{Note:} In order to have more detail about the meaning of the parameters and their default values, please we recommend to check the frameworks mentioned in subsection \ref{frame}.

\begin{table}[H]
	\centering
	\resizebox{\textwidth}{!}{%
		\begin{tabular}{|C{2.8cm}|C{8cm}|C{8cm}|}
			\hline
			\textbf{Stream Learners} & \textbf{Electricity market} & \textbf{Moving Squares} \\ \hline
			\textbf{KNN} & n\_neighbors=$3$, max\_window\_size=$10$, leaf\_size=$2$, pre-training\_size=$11,000$, M\_sw\_size=$500$ & n\_neighbors=$20$, max\_window\_size=$50$, leaf\_size=$2$, pre-training\_size=$12,500$, M\_sw\_size=$500$ \\ \hline
			\textit{\textbf{GRF\_KNN}} & KNN + GRF params (gamma=$2.0$, n\_GRFs=$5$) & KNN + GRF params (gamma=$2.0$, n\_GRFs=$11$) \\ \hline
			\multirow{2}{*}{\textbf{HT}} & parameters by default & parameters by default \\ \cline{2-3} 
			& pre-training\_size=$11,000$, M\_sw\_size=$500$ & pre-training\_size=$12,500$, M\_sw\_size=$500$ \\ \hline
			\textit{\textbf{GRF\_HT}} & HT + GRF params (gamma=$2.0$, n\_GRFs=$5$) & HT + GRF params (gamma=$2.0$, n\_GRFs=$11$) \\ \hline
			\multirow{2}{*}{\textbf{HAT}} & HT + ADWIN params (delta=$0.002$, f=$32$) & HT + ADWIN params (delta=$0.002$, f=$32$) \\ \cline{2-3} 
			& pre-training\_size=$11,000$, M\_sw\_size=$500$ & pre-training\_size=$12,500$, M\_sw\_size=$500$ \\ \hline
			\textit{\textbf{GRF\_HAT}} & HAT + GRF params (gamma=$2.0$, n\_GRFs=$5$) & HAT + GRF params (gamma=$2.0$, n\_GRFs=$11$) \\ \hline
			\multirow{2}{*}{\textbf{MNB}} & alpha=$1.0$, fit\_prior=True & alpha=$1.0$, fit\_prior=True \\ \cline{2-3} 
			& pre-training\_size=$11,000$, M\_sw\_size=$500$ & pre-training\_size=$12,500$, M\_sw\_size=$500$ \\ \hline
			\textit{\textbf{GRF\_MNB}} & MNB + GRF params (gamma=$2.0$, n\_GRFs=$5$) & MNB + GRF params (gamma=$2.0$, n\_GRFs=$11$) \\ \hline
			\multirow{2}{*}{\textbf{GNB}} & var\_smoothing=$1e-9$ & var\_smoothing$=1e-9$ \\ \cline{2-3} 
			& pre-training\_size=$11,000$, M\_sw\_size=$500$ & pre-training\_size=$12,500$, M\_sw\_size=$500$ \\ \hline
			\textit{\textbf{GRF\_GNB}} & GNB + GRF params (gamma=$2.0$, n\_GRFs=$5$) & GNB + GRF params (gamma=$2.0$, n\_GRFs=$11$) \\ \hline
			\multirow{2}{*}{\textbf{SGD}} & parameters by default with n\_iter=$1$ & parameters by default with n\_iter=$1$ \\ \cline{2-3} 
			& pre-training\_size=$11,000$, M\_sw\_size=$500$ & pre-training\_size=$12,500$, M\_sw\_size=$500$ \\ \hline
			\textit{\textbf{GRF\_SGD}} & SGD + GRF params (gamma=$2.0$, n\_GRFs=$5$) & SGD + GRF params (gamma=$2.0$, n\_GRFs=$11$) \\ \hline
			\multirow{2}{*}{\textbf{Perceptron}} & parameters by default with loss='perceptron' and n\_iter=$1$ & parameters by default with loss='perceptron' and n\_iter=$1$ \\ \cline{2-3} 
			& pre-training\_size=$11,000$, M\_sw\_size=$500$ & pre-training\_size=$12,500$, M\_sw\_size=$500$ \\ \hline
			\textit{\textbf{GRF\_Perceptron}} & Perceptron + GRF params (gamma=$2.0$, n\_GRFs=$5$) & Perceptron + GRF params (gamma=$2.0$, n\_GRFs=$11$) \\ \hline
			\multirow{2}{*}{\textbf{PA}} & parameters by default with n\_iter=$1$ & parameters by default with n\_iter=$1$ \\ \cline{2-3} 
			& pre-training\_size=$11,000$, M\_sw\_size=$500$ & pre-training\_size=$12,500$, M\_sw\_size=$500$ \\ \hline
			\textit{\textbf{GRF\_PA}} & PA + GRF params (gamma=$2.0$, n\_GRFs=$5$) & PA + GRF params (gamma=$2.0$, n\_GRFs=$11$) \\ \hline
			\multirow{2}{*}{\textbf{MLP}} & parameters by default with max\_iter=$1$ & parameters by default with max\_iter=$1$ \\ \cline{2-3} 
			& pre-training\_size=$11,000$, M\_sw\_size=$500$ & pre-training\_size=$12,500$, M\_sw\_size=$500$ \\ \hline
			\textit{\textbf{GRF\_MLP}} & MLP + GRF params (gamma=$2.0$, n\_GRFs=$5$) & MLP + GRF params (gamma=$2.0$, n\_GRFs=$11$) \\ \hline
		\end{tabular}	
	}
	\caption{Parameters configuration of the SL methods for datasets \textit{Electricity market} and \textit{Moving squares}.}
	\label{my_parameters_1}
\end{table}

\begin{table}[H]
	\centering
	\resizebox{\textwidth}{!}{%
		\begin{tabular}{|C{2.8cm}|C{8cm}|C{8cm}|}
			\hline
			\textbf{Stream Learners} & \textbf{Weather} & \textbf{SEA} \\ \hline
			\textbf{KNN} & n\_neighbors=$5$, max\_window\_size=$50$, leaf\_size=$2$, pre-training\_size=$4,500$, M\_sw\_size=$500$ & n\_neighbors=$15$, max\_window\_size=$100$, leaf\_size=$2$, pre-training size=$10,000$, M\_SW\_size=$500$ \\ \hline
			\textit{\textbf{GRF\_KNN}} & KNN + GRF params (gamma=$2.0$, n\_GRFs=$3$) & KNN + GRF params (gamma=$2.0$, n\_GRFs=$3$) \\ \hline
			\multirow{2}{*}{\textbf{HT}} & parameters by default & parameters by default \\ \cline{2-3} 
			& pre-training\_size=$4,500$, M\_sw\_size=$500$ & pre-training\_size=$10,000$, M\_sw\_size=$500$ \\ \hline
			\textit{\textbf{GRF\_HT}} & HT + GRF params (gamma=$2.0$, n\_GRFs=$3$) & HT + GRF params (gamma=$2.0$, n\_GRFs=$3$) \\ \hline
			\multirow{2}{*}{\textbf{HAT}} & HT + ADWIN params (delta=$0.002$, f=$32$) & HT + ADWIN params (delta=$0.002$, f=$32$) \\ \cline{2-3} 
			& pre-training\_size=$4,500$, M\_sw\_size=$500$ & pre-training\_size=$10,000$, M\_sw\_size=$500$ \\ \hline
			\textit{\textbf{GRF\_HAT}} & HAT + GRF params (gamma=$2.0$, n\_GRFs=$3$) & HAT + GRF params (gamma=$2.0$, n\_GRFs=$3$) \\ \hline
			\multirow{2}{*}{\textbf{MNB}} & alpha=$1.0$, fit\_prior=True & alpha=$1.0$, fit\_prior=True \\ \cline{2-3} 
			& pre-training\_size=$4,500$, M\_sw\_size=$500$ & pre-training\_size=$10,000$, M\_sw\_size=$500$ \\ \hline
			\textit{\textbf{GRF\_MNB}} & MNB + GRF params (gamma=$2.0$, n\_GRFs=$3$) & MNB + GRF params (gamma=$2.0$, n\_GRFs=$3$) \\ \hline
			\multirow{2}{*}{\textbf{GNB}} & var\_smoothing=$1e-9$ & var\_smoothing=$1e-9$ \\ \cline{2-3} 
			& pre-training\_size=$4,500$, M\_sw\_size=$500$ & pre-training\_size=$10,000$, M\_sw\_size=$500$ \\ \hline
			\textit{\textbf{GRF\_GNB}} & GNB + GRF params (gamma=$2.0$, n\_GRFs=$3$) & GNB + GRF params (gamma=$2.0$, n\_GRFs=$3$) \\ \hline
			\multirow{2}{*}{\textbf{SGD}} & parameters by default with n\_iter=$1$ & parameters by default with n\_iter=$1$ \\ \cline{2-3} 
			& pre-training\_size=$4,500$, M\_sw\_size=$500$ & pre-training\_size=$10,000$, M\_sw\_size=$500$ \\ \hline
			\textit{\textbf{GRF\_SGD}} & SGD + GRF params (gamma=$2.0$, n\_GRFs=$3$) & SGD + GRF params (gamma=$2.0$, n\_GRFs=$3$) \\ \hline
			\multirow{2}{*}{\textbf{Perceptron}} & parameters by default with loss='perceptron' and n\_iter=$1$ & parameters by default with loss='perceptron' and n\_iter=$1$ \\ \cline{2-3} 
			& pre-training\_size=$4,500$, M\_sw\_size=$500$ & pre-training\_size=$10,000$, M\_sw\_size=$500$ \\ \hline
			\textit{\textbf{GRF\_Perceptron}} & Perceptron + GRF params (gamma=$2.0$, n\_GRFs=$3$) & Perceptron + GRF params (gamma=$2.0$, n\_GRFs=$3$) \\ \hline
			\multirow{2}{*}{\textbf{PA}} & parameters by default with n\_iter=$1$ & parameters by default with n\_iter=$1$ \\ \cline{2-3} 
			& pre-training\_size=$4,500$, M\_sw\_size=$500$ & pre-training\_size=$12,500$, M\_sw\_size=$500$ \\ \hline
			\textit{\textbf{GRF\_PA}} & PA + GRF params (gamma=$2.0$, n\_GRFs=$3$) & PA + GRF params (gamma=$2.0$, n\_GRFs=$3$) \\ \hline
			\multirow{2}{*}{\textbf{MLP}} & parameters by default with max\_iter=$1$ & parameters by default with max\_iter=$1$ \\ \cline{2-3} 
			& pre-training\_size=$4,500$, M\_sw\_size=$500$ & pre-training\_size=$12,500$, M\_sw\_size=$500$ \\ \hline
			\textit{\textbf{GRF\_MLP}} & MLP + GRF params (gamma=$2.0$, n\_GRFs=$3$) & MLP + GRF params (gamma=$2.0$, n\_GRFs=$3$) \\ \hline
		\end{tabular}
	}
	\caption{Parameters configuration of the SL methods for datasets \textit{Weather} and \textit{SEA}.}
	\label{my_parameters_2}	
\end{table}

\begin{table}[H]
	\centering
	\resizebox{\textwidth}{!}{%
		\begin{tabular}{|C{2.8cm}|C{8cm}|C{8cm}|}
			\hline
			\textbf{Stream Learners} & \textbf{Airlines} & \textbf{Synthetic datasets} \\ \hline
			\textbf{KNN} & n\_neighbors=$7$, max\_window\_size=$70$, leaf\_size=$2$, pre-training\_size=$12,500$, M\_sw\_size=$500$ & n\_neighbors=$3$, max\_window\_size=$10$, leaf\_size=$2$, pre-training\_size=$12,500$, M\_sw\_size=$500$ \\ \hline
			\textit{\textbf{GRF\_KNN}} & KNN + GRF params (gamma=$2.0$, n\_GRFs=$3$) & KNN + GRF params (gamma=$2.0$, n\_GRFs=$3$) \\ \hline
			\multirow{2}{*}{\textbf{HT}} & parameters by default & parameters by default \\ \cline{2-3} 
			& pre-training\_size=$12,500$, M\_sw\_size=$500$ & pre-training\_size=$12,500$, M\_sw\_size=$500$ \\ \hline
			\textit{\textbf{GRF\_HT}} & HT + GRF params (gamma=$2.0$, n\_GRFs=$3$) & HT + GRF params (gamma=$2.0$, n\_GRFs=$3$) \\ \hline
			\multirow{2}{*}{\textbf{HAT}} & HT + ADWIN params (delta=$0.002$, f=$32$) & HT + ADWIN params (delta=$0.002$, f=$32$) \\ \cline{2-3} 
			& pre-training\_size=$12,500$, M\_sw\_size=$500$ & pre-training\_size=$12,500$, M\_sw\_size=$500$ \\ \hline
			\textit{\textbf{GRF\_HAT}} & HAT + GRF params (gamma=$2.0$, n\_GRFs=$3$) & HAT + GRF params (gamma=$2.0$, n\_GRFs=$3$) \\ \hline
			\multirow{2}{*}{\textbf{MNB}} & alpha=$1.0$, fit\_prior=True & alpha=$1.0$, fit\_prior=True \\ \cline{2-3} 
			& pre-training\_size=$12,500$, M\_sw\_size=$500$ & pre-training\_size=$12,500$, M\_sw\_size=$500$ \\ \hline
			\textit{\textbf{GRF\_MNB}} & MNB + GRF params (gamma=$2.0$, n\_GRFs=$3$) & MNB + GRF params (gamma=$2.0$, n\_GRFs=$3$) \\ \hline
			\multirow{2}{*}{\textbf{GNB}} & var\_smoothing=$1e-9$ & var\_smoothing=$1e-9$ \\ \cline{2-3} 
			& pre-training\_size=$12,500$, M\_sw\_size=$500$ & pre-training\_size=$12,500$, M\_sw\_size=$500$ \\ \hline
			\textit{\textbf{GRF\_GNB}} & GNB + GRF params (gamma=$2.0$, n\_GRFs=$3$) & GNB + GRF params (gamma=$2.0$, n\_GRFs=$3$) \\ \hline
			\multirow{2}{*}{\textbf{SGD}} & parameters by default with n\_iter=$1$ & parameters by default with n\_iter=$1$ \\ \cline{2-3} 
			& pre-training\_size=$12,500$, M\_sw\_size=$500$ & pre-training\_size=$12,500$, M\_sw\_size=$500$ \\ \hline
			\textit{\textbf{GRF\_SGD}} & SGD + GRF params (gamma=$2.0$, n\_GRFs=$3$) & SGD + GRF params (gamma=$2.0$, n\_GRFs=$3$) \\ \hline
			\multirow{2}{*}{\textbf{Perceptron}} & parameters by default with loss='perceptron' and n\_iter=$1$ & parameters by default with loss='perceptron' and n\_iter=$1$ \\ \cline{2-3} 
			& pre-training\_size=$12,500$, M\_sw\_size=$500$ & pre-training\_size=$12,500$, M\_sw\_size=$500$ \\ \hline
			\textit{\textbf{GRF\_Perceptron}} & Perceptron + GRF params (gamma=$2.0$, n\_GRFs=$3$) & Perceptron + GRF params (gamma=$2.0$, n\_GRFs=$3$) \\ \hline
			\multirow{2}{*}{\textbf{PA}} & parameters by default with n\_iter=$1$ & parameters by default with n\_iter=$1$ \\ \cline{2-3} 
			& pre-training\_size=$12,500$, M\_sw\_size=$500$ & pre-training\_size=$12,500$, M\_sw\_size=$500$ \\ \hline
			\textit{\textbf{GRF\_PA}} & PA + GRF params (gamma=$2.0$, n\_GRFs=$3$) & PA + GRF params (gamma=$2.0$, n\_GRFs=$3$) \\ \hline
			\multirow{2}{*}{\textbf{MLP}} & parameters by default with max\_iter=$1$ & parameters by default with max\_iter=$1$ \\ \cline{2-3} 
			& pre-training\_size=$12,500$, M\_sw\_size=$500$ & pre-training\_size=$12,500$, M\_sw\_size=$500$ \\ \hline
			\textit{\textbf{GRF\_MLP}} & MLP + GRF params (gamma=$2.0$, n\_GRFs=$3$) & MLP + GRF params (gamma=$2.0$, n\_GRFs=$3$) \\ \hline
		\end{tabular}
	}
	\caption{Parameters configuration of the SL methods for datasets \textit{Airlines} and \textit{Synthetic datasets} (\textit{circle\_concept1}, \textit{circle\_concept2}, \textit{line\_concept1}, \textit{line\_concept2}, \textit{sine\_concept1}, \textit{sine\_concept2}, \textit{sineH\_concept1}, and \textit{sineH\_concept2}).}
	\label{my_parameters_3}	
\end{table}

\section*{Bibliography}

\bibliographystyle{model5-names}
\bibliography{biblio}

\end{document}